\documentclass[10pt]{article} 
\usepackage[preprint]{tmlr}

\usepackage{amsmath,amsfonts,bm}

\def\eqref#1{equation~\ref{#1}}

\def\1{\bm{1}}

\DeclareMathAlphabet{\mathsfit}{\encodingdefault}{\sfdefault}{m}{sl}
\SetMathAlphabet{\mathsfit}{bold}{\encodingdefault}{\sfdefault}{bx}{n}

\usepackage{hyperref}
\usepackage{subcaption}
\usepackage{cleveref}
\usepackage{url}
\usepackage{graphicx}
\usepackage{booktabs}
\usepackage{multirow}
\usepackage[table,xcdraw]{xcolor}
\usepackage{pifont}
\usepackage{float}
\usepackage{moresize}
\usepackage{setspace}
\usepackage{listings}
\usepackage{xcolor}
\usepackage{xspace}

\crefname{section}{section}{sections}
\crefname{appendix}{appendix}{appendices} 

\newcommand{\eg}{\emph{e.g.,}\xspace}
\newcommand{\ie}{\emph{i.e.}\xspace}

\title{Node-wise Feature Encoding for Neural Performance Prediction}

\author{\name Matthew Grenier$^{1,\dagger}$, \name William Hammer$^{1,*,\dagger}$, \name Andrew Heuer$^{1,*,\dagger}$, \name Nikhil Krishna $^{1,*,\dagger}$, \name Yi Wang$^{2}$, Ramtin Zand$^{1}$ \\ \addr $^{1}$University of South Carolina Department of Computer Science and Engineering, $^{2}$University of South Carolina Department of Mechanical Engineering
 \let\footnotemark\relax\thanks{$^*$ Equal contribution.}
 \let\footnotemark\relax\thanks{$\dagger$ Corresponding authors: mgrenier@email.sc.edu, whammer@email.sc.edu, aheuer@email.sc.edu, nkrishna@email.sc.edu.}
}

\def\month{08}  
\def\year{2026} 
\def\openreview{\url{https://openreview.net/forum?id=XXXX}} 

\begin{document}

\maketitle

\begin{abstract}
As neural networks are increasingly deployed on resource constrained edge devices, accurate prediction of latency and energy is critical for efficient neural architecture search. Existing GNN and transformer based predictors achieve strong results but largely ignore node-level computational cost, limiting their ability to model performance critical operations. To address this, we introduce FeatureFormer, a neural performance predictor that incorporates explicit node-wise encodings of FLOPs, parameter counts, and memory proxies within a gated graph attention architecture. We also present NNEQ, a new large-scale energy consumption dataset that enables unified evaluation of latency and energy prediction. Extensive experiments demonstrate that FeatureFormer achieves state-of-the-art performance across both metrics, including challenging out-of-domain settings. Finally, we show that the proposed encoding is broadly applicable and consistently improves existing predictors with negligible overhead.
\end{abstract}

\section{Introduction}
\label{sec:intro}
Deep neural networks have achieved strong performance across a wide range of domains, particularly in computer vision. However, the growing complexity of modern architectures has made efficient deployment on resource-constrained edge devices increasingly challenging, as applications must satisfy strict constraints on latency, energy consumption, and memory usage. Neural architecture search (NAS) \citep{elsken2019neural,wang2024advances} addresses this challenge by reformulating network design as a search problem, and many hardware-aware variants \citep{cai2018proxylessnas,tan2019mnasnet,wu2019fbnet} explicitly optimize for metrics like latency and energy alongside accuracy, making them well suited to edge deployment.

\begin{figure}[h]
    \centering
    \includegraphics[width=1\linewidth]{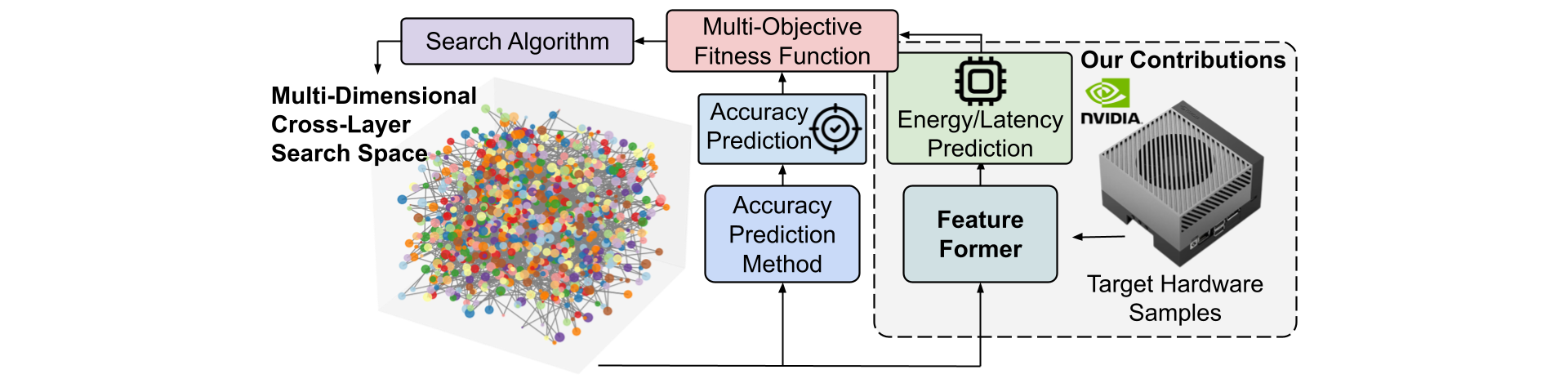}
    \caption{Overview of the proposed neural performance predictor integrated into a neural architecture search (NAS) procedure. By replacing direct measurements of latency and energy with a predictive model, the hardware-aware NAS can be greatly accelerated.}
    \label{fig:nas_overview}
\end{figure}

Because repeated on-device latency and energy measurements make hardware-aware NAS expensive, many frameworks instead train neural performance predictors \citep{cai2017neuralpower,qi2017paleo} that estimate these metrics directly from an architecture's structure (\eg \cref{fig:nas_overview}). Such predictors typically represent networks as directed acyclic graphs, with recent work moving from message-passing GNNs \citep{liu2022nnlqp,dudziak2020brp,panner2023dippm} to graph transformers \citep{yi2023narv1,yi2023narv2,xu2025nn} to better capture topological relationships between operations. While these methods perform well, they approach the problem primarily from a topological perspective, leaving node-level computational information largely unexploited.

\begin{figure}[t]
\centering
\begin{subfigure}{.5\textwidth}
  \centering
  \includegraphics[width=1\linewidth]{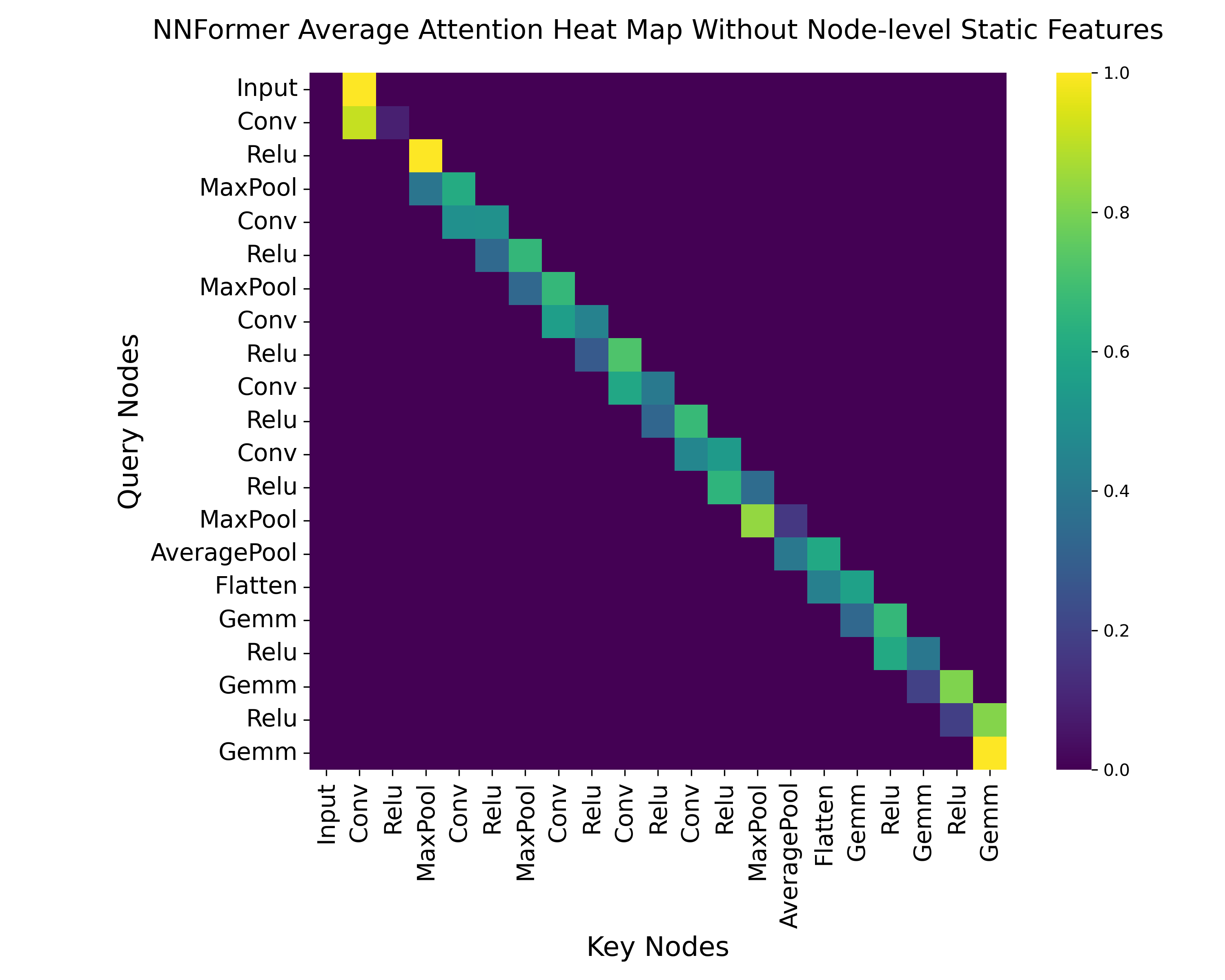}
  \caption{Without our encoding method}
  \label{fig:without_static}
\end{subfigure}%
\begin{subfigure}{.5\textwidth}
  \centering
  \includegraphics[width=1\linewidth]{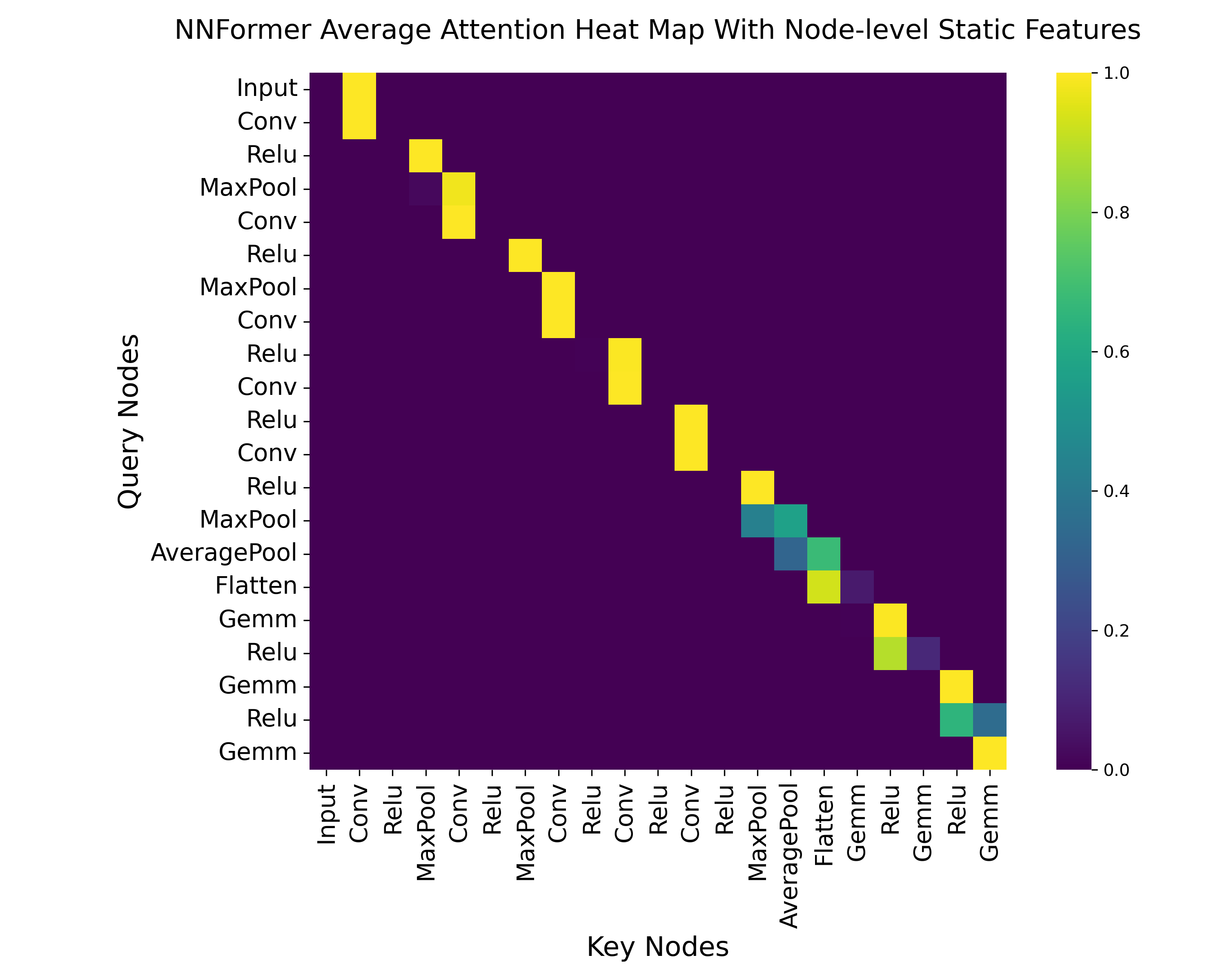}
  \caption{With our encoding method}
  \label{fig:with_static}
\end{subfigure}
\caption{Last-layer attention scores of NN-Former \citep{xu2025nn}, averaged over 200 AlexNet models from the NNLQ dataset \citep{liu2022nnlqp} that share the same architecture. (a) Without the proposed encoding, attention is evenly distributed, often favoring computationally inexpensive operations (\eg ReLU). (b) With the proposed encoding, attention concentrates on operations that are more relevant for latency prediction, such as convolutions.}
\label{fig:attn_map}
\end{figure}

To this end, we argue that a significant amount of highly relevant information inherent in neural architectures remains underutilized. In particular, prior graph-based predictors do not explicitly incorporate layer-specific computational information such as floating point operation (FLOP) counts or parameter sizes at the node-level, so operations that dominate latency or energy consumption are treated similarly to those with negligible cost. As shown in \cref{fig:without_static}, this leaves attention distributed relatively evenly across operations, regardless of whether a node corresponds to a lightweight layer such as ReLU or a computationally intensive one such as convolution.

To address this limitation, we introduce FeatureFormer, a neural architecture performance predictor that incorporates node-level computational information into neural network representations. FeatureFormer augments each node with computational attributes derived from the underlying operation, including FLOP counts and parameter sizes, providing a topology-independent heuristic of a layer's computational cost. As illustrated in \cref{fig:with_static}, this encoding causes attention to concentrate strongly on operations that dominate inference cost, while inexpensive layers receive little emphasis. FeatureFormer further introduces a gating-augmented attention mechanism that improves the effectiveness of graph-based neural architecture predictors. The key contributions of this work are as follows:

\begin{itemize}
    \item We propose FeatureFormer, a new model for neural architecture prediction that combines architectural and encoding innovations to achieve state-of-the-art performance for inference latency and energy consumption prediction.
    \item We introduce NNEQ, an expanded version of the NNLQ \citep{liu2022nnlqp} dataset, that includes energy consumption measurements, providing a large and easily adoptable dataset for training neural predictors.
    \item We provide detailed analyses demonstrating that the proposed encoding approach is applicable to existing methods, offering a means to improve prediction accuracy with negligible trade-offs.
\end{itemize}

\section{Related Work}

\subsection{Neural Architecture Performance Prediction}
Neural architecture performance prediction aims to estimate attributes such as inference latency or energy consumption to accelerate neural architecture search. Early approaches relied on lightweight models \citep{ni2022online,bouzidi2020performance,cai2017neuralpower}, lookup tables \citep{cai2019once,wu2019fbnet,cai2018proxylessnas,yang2018netadapt}, or simple performance proxies \citep{yu2020bignas,qi2017paleo,juneja2025accelerating}. NeuralPower \citep{cai2017neuralpower}, for example, uses polynomial regression to predict latency and power at the layer level, while Paleo \citep{qi2017paleo} estimates runtime using zero cost proxies such as FLOP count combined with hardware characteristics. More recent proxy-based methods improve upon these approaches by modeling performance at the level of execution kernels rather than individual layers. Kernel-based predictors such as NN-Meter \citep{zhang2021nn} or that of \citet{tu2023unveiling} leverage fused operation measurements collected across hardware platforms to achieve higher accuracy. Similar techniques have also been applied to specialized accelerators like TPUs \citep{kaufman2019learned}. However, these methods require extensive profiling and inaccurately simplify the problem (\eg assuming total latency is equal to the sum of kernel latency), motivating the use of graph-based predictors that explicitly model architectural structure.

\subsection{GNN-based predictors}
To address the limitations of kernel based approaches, GNN-based predictors \citep{zhou2024hgnas,panner2023dippm,dudziak2020brp,liu2022nnlqp,kaufman2021learned} have become the dominant paradigm for latency and energy prediction. GNNs, particularly message passing models such as Graph Convolutional Networks (GCNs) \citep{kipf2016semi}, operate on graph structured data by aggregating information from neighboring nodes. Early methods such as BRP-NAS \citep{dudziak2020brp} demonstrated the effectiveness of GCN based predictors for joint accuracy and latency estimation within NAS. By modeling graph topology, these methods achieve significantly higher accuracy than approaches based solely on zero cost proxies such as FLOPs. Building on this idea, NNLQP \citep{liu2022nnlqp} introduced a GraphSAGE \citep{hamilton2017inductive} based predictor that incorporates additional layer level features, including kernel size and stride, for CNN latency prediction. To further enrich the representation, NNLQP appended a global feature vector containing information such as FLOPs and parameter count to the aggregated output of the GraphSAGE encoder before regression. NNLQP’s encoding established the foundation for using zero cost proxies to augment graph-based predictions, but it, and the works that follow \citep{xu2025nn,yi2023narv1,yi2023narv2}, do not integrate layer level FLOPs and parameters directly into node representations. 


\subsection{Transformer Based Predictors}
Recent approaches for neural architecture performance prediction adopt graph transformer-based models \citep{yun2019graph,luo2023transformers}. Graph transformers are appealing due to their attention mechanisms and global receptive field, which allow them to capture long range dependencies. However, a fully global view has been shown to be suboptimal for latency prediction, as nodes that are far apart in the graph often have little relevance to each other in terms of execution time. This limitation is evident in NAR-Former \citep{yi2023narv1}, which performs well for accuracy prediction but fails to outperform earlier methods such as NNLQP \citep{liu2022nnlqp} for latency prediction. 

To mitigate this issue, NAR-Former V2 \citep{yi2023narv2} introduced masked attention based on the adjacency matrix to suppress interactions between distant nodes. Further extending this idea, NN-Former \citep{xu2025nn} incorporated masking between parallel nodes, as such relationships are important for latency prediction due to the potential for parallel execution. While these methods improve topological modeling, they largely retain the encoding strategy introduced by NNLQP with only minor modifications. As a result, individual nodes still lack explicit information about their computational cost. In this work, we address this limitation by adding computational cost directly into the node features, complementing the topological information. 
 
\begin{figure}[!t]
    \centering
    \includegraphics[width=1\linewidth]{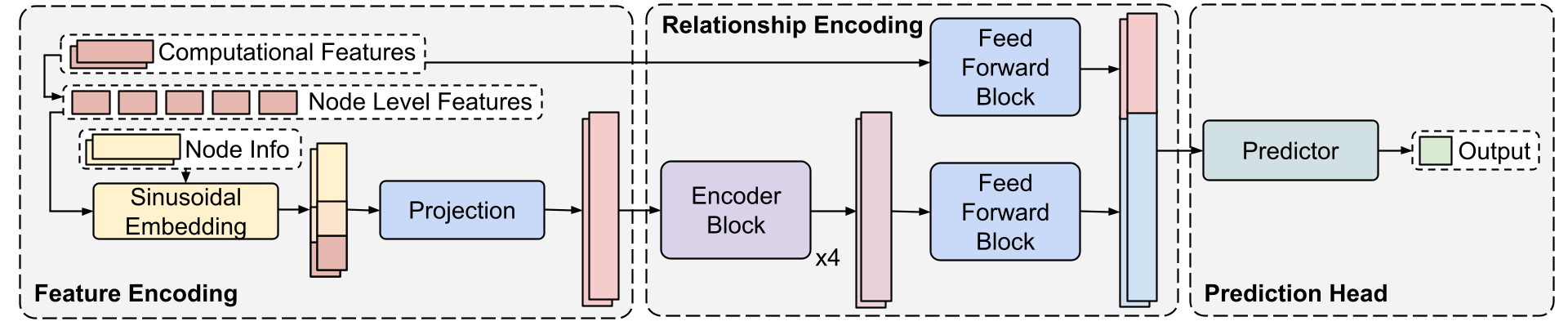}
    \caption{Overview of the proposed FeatureFormer architecture, utilizing global and node-level computational features at different stages of the model.}
    \label{fig:overview}
\end{figure}

\section{Methods}

\Cref{fig:overview} shows the overall architecture of our FeatureFormer model, which is composed of 
two main components: a feature encoding stage that produces node-level representations, and a relationship encoding stage that uses our gated graph attention mechanism to model interactions between nodes in the input architecture. The resulting node representations are then summed, concatenated with global computational features, and passed through a regression head to produce the final latency or energy prediction.

\subsection{Feature Encoding Stage}
\label{sec:method_encoding}
The goal of the feature encoding stage is to provide the model with sufficient information to accurately identify how each node contributes to the inference latency or energy consumption of a given input architecture.

For each node, we generate a vector consisting of three segments: operation type, node attributes, and node computational information. The operation type segment is a one-hot encoding of the operation performed by the node, such as a convolution or ReLU. The node attribute segment contains general information that may influence runtime. At a minimum, this includes the output shape of each layer. For layers with additional structure, such as convolutions, this segment also includes attributes such as kernel size, stride, and padding. Similar to \citet{liu2022nnlqp}, we encode these attributes using sinusoidal encodings. As shown in \cref{eq:sinusodal}, where $x$ denotes the input value and $N$ is half of the desired embedding length, these encodings can be viewed as a generalization of those proposed in \citet{vaswani2017attention} to arbitrary floating point inputs rather than fixed positional indices. The motivation behind this choice of encoding is similar to that described in \citet{tancik2020fourier}, to map single scalars to a higher dimension and allow the model to learn more complex relationships between them. In fact, this method is effectively a Fourier feature encoding. This specific equation was chosen as the frequencies are $log$-spaced, hence very large ranges of values can be effectively encoded. This is crucial for effectively encoding computational information such as FLOPs or parameters which may range from the thousands to the billions.

{
\begin{equation}
\label{eq:sinusodal}
\mathrm{Emb}(x, N) =
\left[
\sin\!\left(\frac{x}{10000^{\frac{0}{N}}}\right),
\cos\!\left(\frac{x}{10000^{\frac{0}{N}}}\right)
\,\ldots\,
\sin\!\left(\frac{x}{10000^{\frac{N-1}{N}}}\right),
\cos\!\left(\frac{x}{10000^{\frac{N-1}{N}}}\right)
\right]
\end{equation}}

The final segment of the node encoding consists of computational proxies. These include the number of FLOPs, the number of parameters, and the multiplied output shape of the layer plus the layer's parameter count, which serves as a proxy for the number of memory operations performed by the given layer. As with the node attributes, these values are encoded using the sinusoidal encodings defined in \cref{eq:sinusodal}. The final node representation is formed by concatenating the three segments, as shown in \cref{eq:encode}, where $F_{op}$, $F_{attr}$, and $F_{comp}$ denote the operation type encoding, attribute encoding, and computational encoding, respectively.

\begin{equation}
\label{eq:encode}
Encode(x) = Concat(F_{op}, F_{attr}, F_{comp})
\end{equation}

\subsection{Relationship Encoding Stage}

\begin{figure}[!t]
\centering
\includegraphics[width=0.9\linewidth]{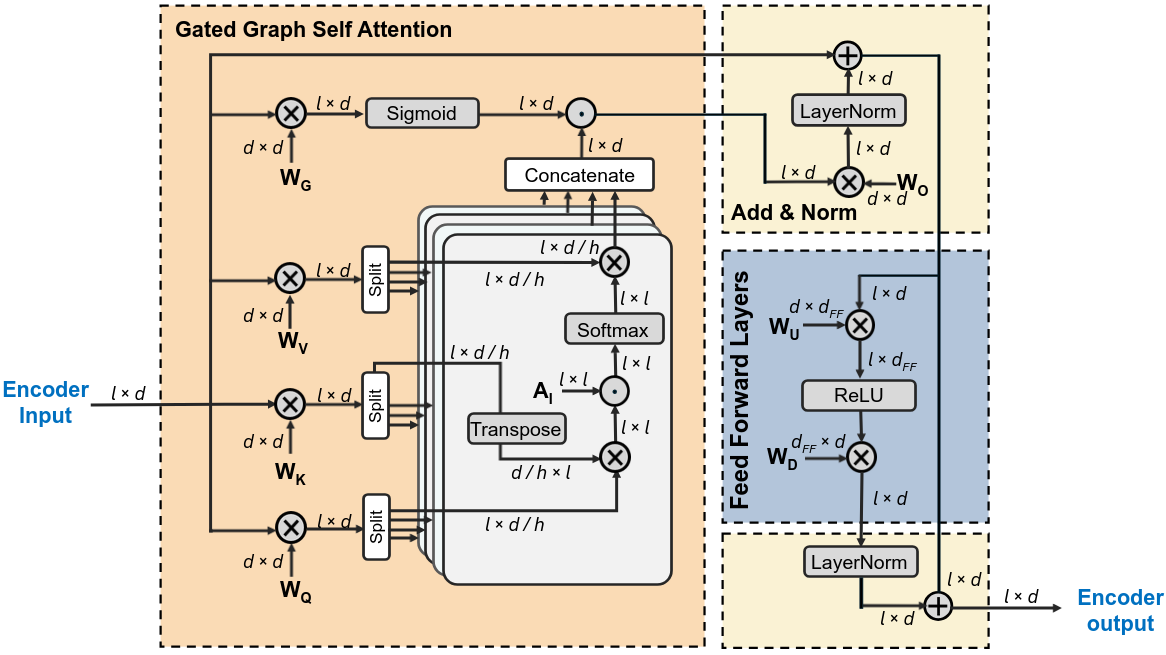}
\caption{Detailed view of an encoder block within our proposed model. In the figure, $A_i$ corresponds to the adjacency based mask used for a given attention head, selected from $A$, $A^T$, $AA^T$, or $A^TA$. For FeatureFormer, we utilize $d = 512$, $d_{FF} = 2048$, and $h = 4$ while $l$ refers to the sequence length.}
\label{fig:graph_attention}
\end{figure}

Given the encoded node representations produced by the feature encoding stage, the relationship encoding stage aims to model how individual operations interact and jointly contribute to the overall latency or energy consumption of the input architecture. The core component of the relationship encoding stage is the encoder block, which is repeated four times to match the number of message passing layers used in NN-Former \citep{xu2025nn}. As illustrated in \cref{fig:graph_attention}, each encoder block consists of a gated graph self attention (GGSA) module followed by a standard two layer feed forward network with ReLU activation.

The GGSA module uses four attention heads, each of which is masked according to a different view of the local graph structure. The first two heads attend over one hop adjacency, using masks $A$ and $A^T$ to capture forward and backward connections, respectively. Following NN-Former \citep{xu2025nn}, the remaining two heads attend over sibling relationships. These correspond to nodes that share a common parent, represented by $A^TA$, and nodes that share a common child, represented by $AA^T$. Such sibling relationships help identify operations that are capable of being executed in parallel. This masking approach is supported by the empirical findings in \citep{xu2025nn}. 

Since each attention head operates over a different adjacency view, we add a gating mechanism so the model can combine their outputs nonlinearly, selectively emphasizing or suppressing a given view of the network. The gating function, shown in \cref{eq:gating}, (where $W^G \in \mathbb{R}^{d \times d}$) follows a multiplicative sigmoid ($\sigma$) formulation, consistent with the analysis presented in \citet{qiu2025gated}. 

\begin{equation}
\label{eq:gating}
G = \sigma(W^GX)
\end{equation}

The full GGSA module is defined as:

\begin{equation}
\label{eq:ggsa}
GGSA(X, A) = (Concat(H_1, H_2, H_3, H_4) \odot G)W^O
\end{equation}

\noindent where the computation of each head's output $H_1 ... H_4$ is described by

\begin{equation}
H_i = softmax \left(\frac{Q_iK_i^T \odot \beta(I + A_i)}{\sqrt{d_h}}\right)V_i 
\end{equation}

\noindent where $A_i \in \{A, A^T, A^TA, AA^T\}$, $I$ denotes the identity matrix, and $\beta$ is the binarization function. The matrices $Q_i = X_iW_i^Q$, $K_i = X_iW_i^K$, and $V_i = X_iW_i^V$ where $W_i^Q, W_i^K, W_i^V \in \mathbb{R}^{d \times d_{head}}$ correspond to the query, key, and value projections for each attention head.

\section{Experiments}
\label{sec:experiments}
To evaluate the effectiveness of FeatureFormer and our encoding method for real-world latency and energy prediction, we consider two primary scenarios. We first focus on out-of-domain prediction (\cref{sec:latency_results,sec:enery_results}), then focus briefly on in-domain prediction (\cref{sec:in_domain}), highlighting the superiority of FeatureFormer in each setting. Across all configurations, we compare our approach to prior state-of-the-art methods and include additional baselines when available. Full training procedures and hyperparameter details are provided in \cref{sec:exp-details}.

\subsection{Datasets}
\label{sec:exp-details}
\begin{figure}[!t]
\centering
\includegraphics[width=1\linewidth]{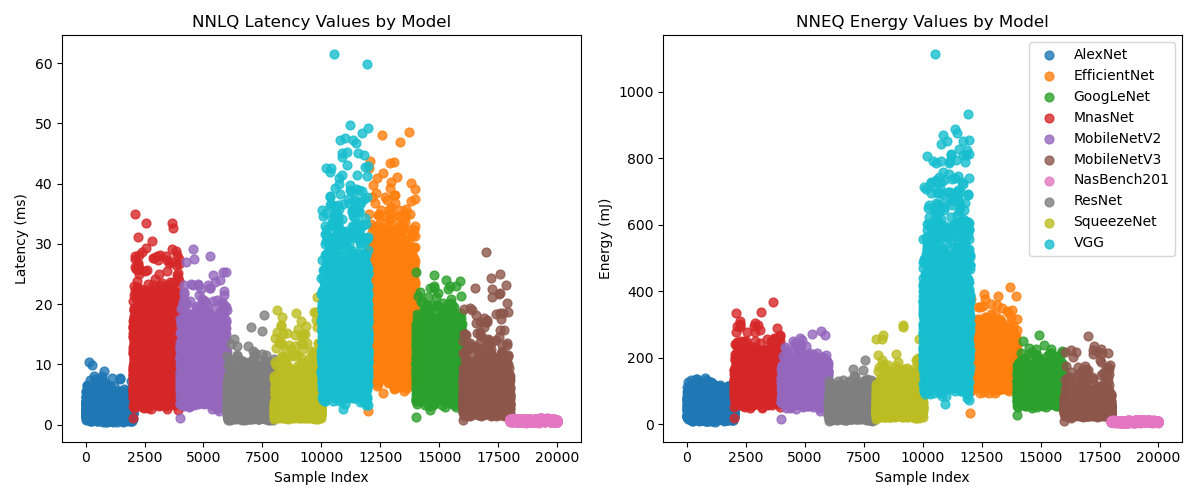}
\caption{NNLQ \citep{liu2022nnlqp} (left) and NNEQ (right) ground truth latency and energy measurements across model families.}
\label{fig:nnlp}
\end{figure}

To benchmark our architecture against prior state-of-the-art latency and energy predictors, we use the standard Neural Network Latency Query (NNLQ) \citep{liu2022nnlqp} dataset, consistent with works such as  \citet{liu2022nnlqp,yi2023narv1,yi2023narv2,xu2025nn}. In addition, to address the lack of a large, open dataset for inference energy prediction, we introduce the Neural Network Energy Query (NNEQ) dataset, a new dataset consisting of energy measurements for various CNNs families. To ensure compatibility with prior work and allow for easy adoption, NNEQ adopts the same dataset structure as by NNLQ, but provides additional measurements recorded on a different hardware platform. This paired design allows performance predictors to be evaluated on both latency and energy prediction, providing a more complete picture of performance across hardware platforms and metrics. Shown in \cref{fig:nnlp}, the distributions of the two datasets vary substantially.

Both datasets include readings from 20,000 models across 10 model families: AlexNet \citep{krizhevsky2012imagenet}, EfficientNet \citep{tan2019efficientnet}, GoogLeNet \citep{szegedy2015going}, MnasNet \citep{tan2019mnasnet}, MobileNetV2 \citep{sandler2018mobilenetv2}, MobileNetV3 \citep{howard2019searching}, NASBench201 \citep{dong2020bench}, ResNet\citep{he2016deep}, SqueezeNet \citep{iandola2016squeezenet}, and VGG16 \citep{simonyan2014very}. Although NNLQ latency measurements were collected on an Nvidia GTX 1660 GPU, we chose to adopt a different, more modern, hardware platform in the Jetson AGX Orin (64 GB) for collecting NNEQ. We note that, although they were not used in this work to maintain continuity with previous studies, latency measurements collected on the Jetson AGX Orin were also included in the NNEQ dataset. As a result, paired latency and energy prediction on the same platform is possible. 

As with NNLQ, all NNEQ models were optimized with TensorRT, with inference energy being measured over a 60 second period. Static features by contrast, such as FLOPs, parameters, and kernel sizes, were precomputed or extracted directly from the model's ONNX graph, in line with prior works \citep{liu2022nnlqp,yi2023narv2,xu2025nn}.

\subsection{Latency Prediction}
\label{sec:latency_results}

Following prior works which utilize out-of-domain prediction \citep{liu2022nnlqp,yi2023narv2,xu2025nn}, we adopt a leave-one-out protocol in which models are trained on nine of the ten model families and evaluated on the held-out family. This procedure is repeated for every held-out family and across ten random seeds. Results are averaged across families to assess a model's ability to generalize to unseen model architectures.

We report two metrics: Mean Absolute Percentage Error (MAPE) and 10\% error bounded accuracy (Acc(10\%)). MAPE captures average predictive error while Acc(10\%) indicates how often predictions fall within 10\% of the ground truth, which is used to indicate a "good enough" prediction. Results are shown in \cref{tab:latency}. FeatureFormer achieves state-of-the-art performance on both metrics, improving over the previous best method, NN-Former \citep{xu2025nn}, by 1.15 percentage points in MAPE and by 2.88 percentage points in Acc(10\%). We also outperform prior methods by a wide margin on several individual families, including NasBench201, MnasNet, and GoogLeNet. Although our performance on the VGG family trails NNLP \citep{liu2022nnlqp} and NAR-Former V2 \citep{yi2023narv2}, we still outperform NN-Former on said family, indicating a more well rounded model. Furthermore, it is clear that NN-Meter \citep{zhang2021nn}, the selected representative of kernel level predictors, is outperformed by the graph-based methods. This observation leads us to focus solely on GNN based methods for the remainder of this work.

\begin{table}
\renewcommand\bfdefault{b}
\scriptsize
\centering
\caption{\textbf{Out-of-domain Latency Prediction on NNLQ \citep{liu2022nnlqp}}. Models are trained on 9 domains and tested on the 10th, hence a "Test Domain" of AlexNet means that the models were trained on all non AlexNet models, with zero shot prediction being performed on AlexNet. When applicable, metrics (MAPE and Acc (10\%)) are reported on the average and best over a set of 10 independent trials.}
\label{tab:latency}
\begin{tabular}{@{}c|l|c c c c c@{}}
\toprule
\multicolumn{1}{c}{Metric} &
\multicolumn{1}{l}{Test Domain} &
\multicolumn{1}{c}{\begin{tabular}[c]{@{}c@{}}
nn-Meter \\ \citet{zhang2021nn} \\
\textit{MobiSys~2021} \\
(avg)
\end{tabular}} &
\multicolumn{1}{c}{\begin{tabular}[c]{@{}c@{}}
NNLP \\ \cite{liu2022nnlqp} \\
\textit{ICPP~2022} \\
(avg / best)
\end{tabular}} &
\multicolumn{1}{c}{\begin{tabular}[c]{@{}c@{}}
NAR-Former V2 \\ \citet{yi2023narv2} \\
\textit{NeurIPS~2023} \\
(avg / best)
\end{tabular}} &
\multicolumn{1}{c}{\begin{tabular}[c]{@{}c@{}}
NN-Former \\ \citet{xu2025nn} \\
\textit{CVPR~2025} \\
(avg / best)
\end{tabular}} &
\multicolumn{1}{c}{\begin{tabular}[c]{@{}c@{}}
FeatureFormer \\
\textit{Ours} \\
(avg / best)
\end{tabular}} \\
\midrule
\multirow{11}{*}{\rotatebox{90}{MAPE $\downarrow$}}
& AlexNet        & \textbf{7.20} & 10.64 / 9.71 & 24.28 / 18.29 & 11.47 / 11.17 & 10.15 / 9.51 \\
& EfficientNet   & 18.93 & 21.46 / 18.72 & 13.20 / 11.37 & 5.13 / 4.81 & \textbf{4.61 / 3.53} \\
& GoogLeNet      & 11.71 & 13.28 / 10.90 & 6.61 / 6.15 & 6.74 / 6.65 & \textbf{4.85 / 4.77} \\
& MnasNet        & 10.69 & 12.07 / 10.86 & 7.16 / 5.93 & 2.71 / 2.54 & \textbf{1.91 / 1.79} \\
& MobileNetV2    & 6.43  & 8.87 / 7.34  & 6.73 / 5.65 & 4.17 / 3.66 & \textbf{2.90 / 2.30} \\
& MobileNetV3    & 35.27 & 14.57 / 13.17 & 9.06 / 8.72 & 9.07 / 9.03 & \textbf{8.01 / 7.71} \\
& NasBench201    & 9.57  & 9.60 / 8.19  & 9.21 / 7.89 & 7.93 / 7.71 & \textbf{6.79 / 4.98} \\
& ResNet         & 15.58 & 7.54 / 7.12  & \textbf{6.80 / 6.44} & 7.49 / 7.38 & 7.98 / 7.10 \\
& SqueezeNet     & 18.69 & 9.84 / 9.52  & 7.08 / 6.56 & 9.08 / 7.05 & \textbf{6.41 / 5.86} \\
& VGG            & 19.47 & \textbf{7.60 / 7.17} & 15.40 / 14.26 & 20.12 / 19.64 & 18.79 / 17.30 \\
\cmidrule{2-7}
& Average        & 15.35 & 11.55 / 10.27 & 10.55 / 9.13 & 8.39 / 7.96 & \textbf{7.24 / 6.48} \\
\midrule
\multirow{11}{*}{\rotatebox{90}{Acc (10\%) $\uparrow$}}
& AlexNet        & \textbf{75.45} & 59.07 / 64.40 & 24.65 / 28.60 & 56.08 / 57.10 & 60.76 / 64.40 \\
& EfficientNet   & 23.40 & 25.37 / 28.80 & 44.01 / 50.20 & \textbf{90.85} / 90.90 & 90.59 / \textbf{95.95} \\
& GoogLeNet      & 47.40 & 36.30 / 48.75 & 80.10 / 83.35 & 80.43 / 83.40 & \textbf{89.71 / 90.55} \\
& MnasNet        & 60.95 & 55.89 / 61.25 & 73.46 / 81.60 & 98.65 / 98.70 & \textbf{99.76 / 99.85} \\
& MobileNetV2    & 80.75 & 63.03 / 72.50 & 78.45 / 83.80 & 94.90 / 96.85 & \textbf{97.31 / 99.65} \\
& MobileNetV3    & 23.45 & 43.26 / 49.65 & 68.43 / 70.50 & \textbf{74.18} / 74.30 & 73.38 / \textbf{75.75} \\
& NasBench201    & 60.65 & 60.70 / 70.60 & 63.13 / 71.70 & 69.90 / 71.10 & \textbf{77.28 / 90.25} \\
& ResNet         & 39.45 & 72.88 / 76.40 & \textbf{77.24 / 79.70} & 70.83 / 71.55 & 69.01 / 75.25 \\
& SqueezeNet     & 36.20 & 58.69 / 60.40 & 75.01 / 79.25 & 77.85 / 80.95 & \textbf{79.05 / 83.55} \\
& VGG            & 26.50 & \textbf{71.04 / 73.75} & 45.21 / 45.30 & 29.40 / 29.85 & 35.04 / 39.25 \\
\cmidrule{2-7}
& Average        & 47.42 & 54.62 / 60.65 & 62.97 / 67.40 & 74.31 / 75.47 & \textbf{77.19 / 81.45} \\
\bottomrule
\end{tabular}
\end{table}

\subsection{Energy Prediction}
\label{sec:enery_results}

Since NNEQ matches NNLQ \citep{liu2022nnlqp} in model content, we use the same metrics, training protocol, and hyperparameters as in the latency experiments. Results are shown in \cref{tab:energy}. Mirroring latency prediction, FeatureFormer achieves state-of-the-art accuracy for energy prediction, with an average MAPE improvement of 0.85 percentage points and a 3.93 percentage point gain in Acc(10\%) over the next best method. Notably, the ranking of competing methods differs from the latency case: NNLP \citep{liu2022nnlqp} outperforms NN-Former \citep{xu2025nn} in MAPE on NNEQ, while NN-Former retains an advantage in Acc(10\%). This pattern suggests that GNN-based methods like NNLP can achieve high average accuracy, but may be less consistent than transformer-based models such as NN-Former in producing predictions within tight error bounds.

\begin{table}[t]
\renewcommand\bfdefault{b}
\scriptsize
\centering
\caption{\textbf{Out-of-domain Energy Prediction on NNEQ.} The testing procedure is shared with that of \cref{tab:latency}}
\label{tab:energy}
\begin{tabular}{@{}cllllclclclcl@{}}
\toprule
Metric &  & Test Domain &  &  & \begin{tabular}[c]{@{}c@{}}NNLP \\\cite{liu2022nnlqp}\\ (avg / best)\end{tabular} &  & \begin{tabular}[c]{@{}c@{}}NAR-Former V2 \\\cite{yi2023narv2}\\ (avg / best)\end{tabular} &  & \begin{tabular}[c]{@{}c@{}}NN-Former \\\cite{xu2025nn}\\ (avg / best)\end{tabular} &  & \begin{tabular}[c]{@{}c@{}}FeatureFormer\\ (avg / best)\end{tabular} &  \\ \midrule
\multicolumn{1}{c|}{\multirow{11}{*}{\rotatebox{90}{MAPE $\downarrow$}}} &  & AlexNet & \multicolumn{1}{l|}{} &  & 10.70 / 8.06 &  & 16.68 / 12.44 &  & 7.74 / 5.94 &  & \textbf{7.29 / 5.42} &  \\
\multicolumn{1}{c|}{} &  & EfficientNet & \multicolumn{1}{l|}{} &  & 6.45 / 6.02 &  & 4.49 / 4.09 &  & 3.24 / 2.80 &  & \textbf{2.89 / 2.47} &  \\
\multicolumn{1}{c|}{} &  & GoogLeNet & \multicolumn{1}{l|}{} &  & 3.89 / 3.49 &  & \textbf{3.44 / 3.31} &  & 3.83 / 3.45 &  & 3.63 / 3.35 &  \\
\multicolumn{1}{c|}{} &  & MnasNet & \multicolumn{1}{l|}{} &  & 9.18 / 7.81 &  & 4.49 / 4.01 &  & 3.28 / 2.72 &  & \textbf{3.10 / 2.72} &  \\
\multicolumn{1}{c|}{} &  & MobileNetV2 & \multicolumn{1}{l|}{} &  & 5.98 / 4.94 &  & 3.78 / 3.42 &  & 3.34 / 2.93 &  & \textbf{3.04 / 2.70} &  \\
\multicolumn{1}{c|}{} &  & MobileNetV3 & \multicolumn{1}{l|}{} &  & 13.75 / 7.36 &  & 8.23 / 7.26 &  & \textbf{4.59 / 4.36} &  & 5.18 / 4.79 &  \\
\multicolumn{1}{c|}{} &  & NasBench201 & \multicolumn{1}{l|}{} &  & 10.80 / 8.36 &  & 11.23 / 9.09 &  & 15.67 / \textbf{6.15} &  & \textbf{8.94} / 7.45 &  \\
\multicolumn{1}{c|}{} &  & ResNet & \multicolumn{1}{l|}{} &  & 5.69 / 5.45 &  & 5.31 / 5.03 &  & 5.00 / 4.62 &  & \textbf{4.65 / 4.36} &  \\
\multicolumn{1}{c|}{} &  & SqueezeNet & \multicolumn{1}{l|}{} &  & 6.96 / 6.34 &  & 6.11 / 5.47 &  & 5.20 / 4.75 &  & \textbf{4.57 / 4.15} &  \\
\multicolumn{1}{c|}{} &  & VGG & \multicolumn{1}{l|}{} &  & \textbf{4.71 / 4.53} &  & 15.98 / 14.93 &  & 26.47 / 25.49 &  & 23.05 / 21.17 &  \\ \cmidrule{2-13} 
\multicolumn{1}{c|}{} &  & Average & \multicolumn{1}{l|}{} &  & 7.81 / 6.24 &  & 7.97 / 6.91 &  & 7.84 / 6.32 &  & \textbf{6.63 / 5.86} &  \\ \midrule
\multicolumn{1}{c|}{\multirow{11}{*}{\rotatebox{90}{Acc (10\%) $\uparrow$}}} &  & AlexNet & \multicolumn{1}{l|}{} &  & 60.85 / 72.25 &  & 43.73 / 63.40 &  & 69.95 / 80.05 &  & \textbf{73.52 / 85.80} &  \\
\multicolumn{1}{c|}{} &  & EfficientNet & \multicolumn{1}{l|}{} &  & 78.48 / 82.40 &  & 92.11 / 94.75 &  & 97.90 / 99.10 &  & \textbf{98.62 / 99.75} &  \\
\multicolumn{1}{c|}{} &  & GoogLeNet & \multicolumn{1}{l|}{} &  & 96.03 / 97.55 &  & \textbf{97.32 / 97.95} &  & 95.40 / 97.35 &  & 96.62 / 97.95 &  \\
\multicolumn{1}{c|}{} &  & MnasNet & \multicolumn{1}{l|}{} &  & 58.99 / 69.95 &  & 92.37 / 95.40 &  & 98.04 / 99.20 &  & \textbf{98.76 / 99.75} &  \\
\multicolumn{1}{c|}{} &  & MobileNetV2 & \multicolumn{1}{l|}{} &  & 81.72 / 88.15 &  & 95.93 / 97.40 &  & 97.62 / 98.75 &  & \textbf{98.48 / 99.15} &  \\
\multicolumn{1}{c|}{} &  & MobileNetV3 & \multicolumn{1}{l|}{} &  & 36.10 / 72.60 &  & 68.45 / 74.35 &  & \textbf{91.16 / 92.30} &  & 87.67 / 89.70 &  \\
\multicolumn{1}{c|}{} &  & NasBench201 & \multicolumn{1}{l|}{} &  & 55.17 / 66.55 &  & 55.43 / 65.35 &  & 45.37 / \textbf{79.95} &  & \textbf{64.61} / 73.90 &  \\
\multicolumn{1}{c|}{} &  & ResNet & \multicolumn{1}{l|}{} &  & 84.58 / 86.10 &  & 86.90 / 89.45 &  & 89.04 / 91.70 &  & \textbf{91.37 / 93.30} &  \\
\multicolumn{1}{c|}{} &  & SqueezeNet & \multicolumn{1}{l|}{} &  & 75.43 / 79.75 &  & 80.61 / 85.20 &  & 87.43 / 90.25 &  & \textbf{91.56 / 94.35} &  \\
\multicolumn{1}{c|}{} &  & VGG & \multicolumn{1}{l|}{} &  & \textbf{89.89 / 90.85} &  & 45.31 / 52.35 &  & 13.12 / 19.10 &  & 23.10 / 28.15 &  \\ \cmidrule{2-13} 
\multicolumn{1}{c|}{} &  & Average & \multicolumn{1}{l|}{} &  & 71.72 / 80.62 &  & 75.81 / 81.56 &  & 78.50 / 84.78 &  & \textbf{82.43 / 86.18} &  \\ \bottomrule
\end{tabular}
\end{table}

\begin{table}[h]
\caption{\textbf{In-domain latency and energy prediction on NNLQ \citep{liu2022nnlqp} and NNEQ.} As in \cref{tab:latency,tab:energy}, each result is reported over 10 independent experiments.}
\label{tab:in_domain}
\centering
\renewcommand\bfdefault{b}
\begin{tabular}{@{}l lcccc@{}}
\toprule
\textbf{Dataset} & \textbf{Metric}
& \begin{tabular}[c]{@{}c@{}}NNLP \\\cite{liu2022nnlqp}\\ (avg / best)\end{tabular}
& \begin{tabular}[c]{@{}c@{}}NAR-Former V2 \\\cite{yi2023narv2}\\ (avg / best)\end{tabular}
& \begin{tabular}[c]{@{}c@{}}NN-Former \\\cite{xu2025nn}\\ (avg / best)\end{tabular}
& \begin{tabular}[c]{@{}c@{}}FeatureFormer\\ (avg / best)\end{tabular} \\
\midrule

\multirow{2}{*}{NNLQ}
& MAPE $\downarrow$
& 3.47 / 3.44
& 3.07 / 3.00
& 2.85 / 2.65
& \textbf{2.57 / 2.46} \\

& Acc (10\%) $\uparrow$
& 95.25 / 95.50
& 96.41 / 96.30
& 97.45 / 97.85
& \textbf{97.58 / 97.95} \\

\midrule

\multirow{2}{*}{NNEQ}
& MAPE $\downarrow$
& 3.70 / 3.57
& 3.02 / 2.89
& 2.77 / 2.50
& \textbf{2.14 / 2.06} \\

& Acc (10\%) $\uparrow$
& 95.89 / 97.05
& 97.91 / 98.45
& 98.46 / 99.30
& \textbf{99.31 / 99.60} \\

\bottomrule
\end{tabular}
\end{table}

\subsection{In-Domain Prediction}
\label{sec:in_domain}
Following \citet{yi2023narv2,xu2025nn}, we additionally conducted experiments on in-domain prediction, the case where the models are trained and tested on the same distribution. To do this, from each model family, 1800 models were randomly selected for training, leaving the remaining 200 for testing, totaling 18000 train and 2000 test models. The results for these experiments are shown in \cref{tab:in_domain}. For latency prediction on NNLQ \citep{liu2022nnlqp}, we outperform the previous state-of-the-art model NN-Former \citep{xu2025nn} by 0.28 MAPE and Acc(10\%) of 0.13 on average. Additionally, for the case of energy prediction on NNEQ, we once again substantially outperform NN-Former by 0.63 MAPE and 0.85 Acc(10\%). Notably, our \textit{average} performance on NNEQ across 10 seeds is better than the \textit{best} performance of any other models, highlighting the effectiveness of our method.

\subsection{Encoding Experiments}
\label{sec:encoding_exp}

Beyond evaluating our full model, we test the impact of our encoding by adding $F_{comp}$ to several prior models that also employ NNLP-style encodings. These models all make use of some level of $F_{op}$ and $F_{attr}$, although the exact makeup and relative length of each segment differs. Due to the possibility that the introduction of $F_{comp}$ could change the optimal length of $F_{op}$ and $F_{attr}$, we conducted a small grid search for each model, aiming to discern the optimal length $F_{op}$ and $F_{attr}$ when coupled with $F_{comp}$. For simplicity, we kept the length of $F_{comp}$ fixed between all models. Tables \ref{tab:enc_ablation_latency} and \ref{tab:enc_ablation_energy} summarize these results by reporting the average MAPE/Acc(10\%) on NNLQ \citep{liu2022nnlqp} and NNEQ, respectively. Adding $F_{comp}$ substantially improves MAPE and Acc(10\%) for almost every model and task. For example, NNLP gains 8.04 percentage points in average Acc(10\%) on NNLQ after the introduction of $F_{comp}$. The only exception is a slight drop in average MAPE and Acc(10\%) for NN-Former on NNEQ as shown in \Cref{tab:enc_ablation_energy}.

\begin{table}[h]
\caption{\textbf{Performance of various models on NNLQ \citep{liu2022nnlqp} (out-of-domain) with node-level computational information ($F_{comp}$) added.} Due to space constraints, results for individual model families are omitted here. Please refer to the appendix for full results.}
\label{tab:enc_ablation_latency}
\centering
\renewcommand\bfdefault{b}
\begin{tabular}{@{}llcccccccl@{}}
\toprule
\multicolumn{1}{c}{Metric} &  & \begin{tabular}[c]{@{}c@{}}NNLP \\\cite{liu2022nnlqp}\\ (avg / best)\end{tabular} & \multicolumn{1}{l}{} & \begin{tabular}[c]{@{}c@{}}NAR-Former V2 \\\cite{yi2023narv2}\\ (avg / best)\end{tabular} & \multicolumn{1}{l}{} & \begin{tabular}[c]{@{}c@{}}NN-Former \\\cite{xu2025nn}\\ (avg / best)\end{tabular} & \multicolumn{1}{l}{} & \begin{tabular}[c]{@{}c@{}}FeatureFormer\\ (avg / best)\end{tabular} &  \\ \midrule
MAPE $\downarrow$ &  & \begin{tabular}[c]{@{}c@{}}10.07 / 8.11\\ ({\color[HTML]{349C07}-1.48} / {\color[HTML]{349C07}-2.16})\end{tabular} &  & \begin{tabular}[c]{@{}c@{}}8.94 / 7.79\\ ({\color[HTML]{349C07}-1.61} / {\color[HTML]{349C07}-1.33})\end{tabular} &  & \begin{tabular}[c]{@{}c@{}}7.60 / 6.94\\ ({\color[HTML]{349C07}-0.79} / {\color[HTML]{349C07}-1.02})\end{tabular} &  & \begin{tabular}[c]{@{}c@{}}\textbf{7.24} / \textbf{6.48}\\ ({\color[HTML]{349C07}-0.39} / {\color[HTML]{349C07}-0.23})\end{tabular} &  \\[7pt]
Acc (10\%) $\uparrow$ &  & \begin{tabular}[c]{@{}c@{}}62.66 / 71.05\\ ({\color[HTML]{349C07}+8.03} / {\color[HTML]{349C07}+10.40})\end{tabular} &  & \begin{tabular}[c]{@{}c@{}}69.43 / 75.46\\ ({\color[HTML]{349C07}+6.46} / {\color[HTML]{349C07}+8.06})\end{tabular} &  & \begin{tabular}[c]{@{}c@{}}75.70 / 79.84\\ ({\color[HTML]{349C07}+1.39} / {\color[HTML]{349C07}+4.37})\end{tabular} &  & \begin{tabular}[c]{@{}c@{}}\textbf{77.19} / \textbf{81.45}\\ ({\color[HTML]{349C07}+2.54} / {\color[HTML]{349C07}+1.48})\end{tabular} &  \\ \bottomrule
\end{tabular}
\end{table}

\begin{table}[h]
\caption{Performance of various models on NNEQ (out-of-domain) with node-level computational information ($F_{comp}$) added.}
\label{tab:enc_ablation_energy}
\renewcommand\bfdefault{b}
\centering
\begin{tabular}{@{}llcccccccl@{}}
\toprule
\multicolumn{1}{c}{Metric} &  & \begin{tabular}[c]{@{}c@{}}NNLP \\\cite{liu2022nnlqp}\\ (avg / best)\end{tabular} & \multicolumn{1}{l}{} & \begin{tabular}[c]{@{}c@{}}NAR-Former V2 \\\cite{yi2023narv2}\\ (avg / best)\end{tabular} & \multicolumn{1}{l}{} & \begin{tabular}[c]{@{}c@{}}NN-Former \\\cite{xu2025nn}\\ (avg / best)\end{tabular} & \multicolumn{1}{l}{} & \begin{tabular}[c]{@{}c@{}}FeatureFormer\\ (avg / best)\end{tabular} &  \\ \midrule
MAPE $\downarrow$ &  & \begin{tabular}[c]{@{}c@{}}7.48 / 6.05\\ ({\color[HTML]{349C07}-0.33} / {\color[HTML]{349C07}-0.18})\end{tabular} &  & \begin{tabular}[c]{@{}c@{}}6.88 / 5.88\\ ({\color[HTML]{349C07}-1.09} / {\color[HTML]{349C07}-1.03})\end{tabular} &  & \begin{tabular}[c]{@{}c@{}}7.88 / 6.17\\ ({\color[HTML]{C82F09}+0.04} / {\color[HTML]{349C07}-0.15})\end{tabular} &  & \begin{tabular}[c]{@{}c@{}}\textbf{6.63} / \textbf{5.86}\\ ({\color[HTML]{349C07}-0.17} / {\color[HTML]{349C07}-0.20})\end{tabular} &  \\[7pt]
Acc (10\%) $\uparrow$ &  & \begin{tabular}[c]{@{}c@{}}73.97 / 82.46\\ ({\color[HTML]{349C07}+2.25} / {\color[HTML]{349C07}+1.85})\end{tabular} &  & \begin{tabular}[c]{@{}c@{}}81.29 / \textbf{86.56}\\ ({\color[HTML]{349C07}+5.48} / {\color[HTML]{349C07}+5.00})\end{tabular} &  & \begin{tabular}[c]{@{}c@{}}78.25 / 86.01\\ ({\color[HTML]{C82F09}-0.25} / {\color[HTML]{349C07}+1.24})\end{tabular} &  & \begin{tabular}[c]{@{}c@{}}\textbf{82.43} / 86.18\\ ({\color[HTML]{349C07}+0.63} / {\color[HTML]{349C07}+0.45})\end{tabular} &  \\ \bottomrule
\end{tabular}
\end{table}

Aside from this exception, the improvement holds consistently across models and tasks, indicating that node-level computational encoding is broadly applicable to neural performance prediction, not specific to our architecture. With this being said, although prior models benefit from the added features, FeatureFormer still attains the best overall performance. We attribute this to its closer adherence to the standard transformer design. NAR-Former V2 \citep{yi2023narv2} includes attention and feed-forward components, but its attention mechanism is non-standard: it shares a single linear projection for queries and keys and reuses the untransformed input as values. NN-Former \citep{xu2025nn} includes a conventional attention block but also injects message passing into its feed-forward blocks via a simple GCN. FeatureFormer instead uses attention as its only message-passing mechanism and adds a multiplicative gating layer between $W^V$ and $W^O$. We demonstrate the effect of gating in \cref{sec:gating}, and isolate the effect of these architectural changes alone against NN-Former in \cref{sec:scaling}.

\section{Ablations}

In this section we report ablation studies on NNLQ \citep{liu2022nnlqp} and NNEQ that probe the contributions of key model components. We should note that due to quantity of computation required to run a full 10 seed experiment, \ie training a given model 100 times, all ablations (with the exception of \cref{tab:gating_energy}) were run on 3 constant seeds instead of 10, hence results for FeatureFormer may differ slightly from those in \cref{sec:experiments}.

\begin{table}[h]
\caption{\textbf{Effect of gating on FeatureFormer.} Results are reported on the NNEQ dataset (out-of-domain), with average and best reported over 10 independent trials.}
\label{tab:gating_energy}
\renewcommand\bfdefault{b}
\centering
\begin{tabular}{@{}llclc@{}}
\toprule
\multicolumn{1}{c}{Metric} &  & \begin{tabular}[c]{@{}c@{}}No Gating\\ (avg / best)\end{tabular} &  & \begin{tabular}[c]{@{}c@{}}Gating\\ (avg / best)\end{tabular} \\ \midrule
MAPE $\downarrow$                      &  & 7.04 / 6.08                                                      &  & \textbf{6.63 / 5.86}                                          \\[3pt]
Acc (10\%) $\uparrow$                &  & 80.66 / 84.87                                                    &  & \textbf{82.43 / 86.18}                                        \\ \bottomrule
\end{tabular}
\end{table}

\subsection{Effect of Gating}
\label{sec:gating}
To measure the benefit of the gating layer in GGSA, we remove the gating term $G$ from \cref{eq:ggsa} and retrain the model on NNEQ. \Cref{tab:gating_energy} shows that removing the gating term increases MAPE by 0.41 percentage points and reduces Acc(10\%) by 1.77 percentage points. This indicates that the multiplicative gate provides a consistent improvement in both average accuracy and prediction reliability.

\begin{table}[h]
\caption{\textbf{Effect of different masking schemes on FeatureFormer}. Results are reported on the NNEQ dataset (out-of-domain), utilizing 4 attention heads in each configuration.}
\label{tab:mask_ablation}

\renewcommand\bfdefault{b}
\centering
\begin{tabular}{llcccccl}
\toprule
\multicolumn{1}{c}{Metric (avg / best)} &  & Global & \multicolumn{1}{l}{} & $A + A^T + A^TA + AA^T$ & \multicolumn{1}{l}{} & $A, A^T, A^TA, AA^T$ &  \\ \midrule
MAPE $\downarrow$     &  & 8.45 / 7.92 &  & 6.92 / 6.40 &  & \textbf{6.28 / 5.92} &  \\[3pt]
Acc (10\%) $\uparrow$ &  & 75.86 / 78.49 &  & 81.13 / 83.67 &  & \textbf{84.12 / 86.08 } &  \\ \hline
\end{tabular}
\end{table}

\subsection{Masking Variations}
We also study how different masking strategies affect performance. To do this, we apply different masking schemes to FeatureFormer, utilizing four attention heads in all configurations. Shown in \cref{tab:mask_ablation}, we investigate three configurations: global attention, shared masking ($A + A^T + A^TA + AA^T$) where each head is masked the same, and specialized masking ($A, A^T, A^TA, AA^T$) where each head receives a different attention mask. Shown in \cref{tab:mask_ablation}, global attention consistently underperforms, in line with previous works \citep{yi2023narv2,xu2025nn}, hence gating does not overcome the difficulties of global masking. Furthermore, shared masking was outperformed by specialized masking, supporting the claim that separate masking for each head allows for greater head-wise specialization.  

\subsection{Encoding Composition}
\label{sec:ab-encoding}
Here, we investigate the importance and contribution of each of the three major components of our encoding, $F_{op}$, $F_{attr}$, and $F_{comp}$, shown in \cref{tab:enc_segments}. Our results show that each component on their own do not achieve satisfactory results, hence the importance of combining these segments. On average, the best performing configuration was (3), corresponding to an encoding containing solely $F_{attr}$ and $F_{comp}$, with (7), the encoding consisting of all segments, coming in as a close second. While slightly superior to (7) on average, this is mainly due to a vast improvement solely on the NasBench201 family. In fact, (7) out performs (3) on effectively all families except NasBench201 (see \cref{tab:app_enc}), hence we believe that operation type information, specified by $F_{op}$, is worthwhile in practice as it leads to a more generalized model, despite a slight decrease in average accuracy in this instance.

\newcommand{\cmark}{\textcolor{green}{\ding{51}}}
\newcommand{\xmark}{\textcolor{red}{\ding{55}}}

\begin{table}[ht]
\centering
\renewcommand\bfdefault{b}
\caption{Importance of $F_{op}$, $F_{attr}$, and $F_{comp}$ to FeatureFormer on NNLQ \citep{liu2022nnlqp} (out-of-domain).}
\label{tab:enc_segments}
\setlength{\tabcolsep}{8pt}
\begin{tabular}{c c c c c c}
\toprule
\multirow{2}{*{\#}} &
\multicolumn{3}{c}{Components} &
\multicolumn{2}{c}{Metrics (avg / best)} \\
\cmidrule(lr){2-4} \cmidrule(lr){5-6}
 & $F_{op}$ & $F_{attr}$ & $F_{comp}$ & MAPE $\downarrow$ & Acc (10\%) $\uparrow$ \\
\midrule
(1) & \xmark & \xmark & \cmark & 19.77 / 18.49 & 33.93 / 38.23 \\
(2) & \xmark & \cmark & \xmark & 8.18 / 7.66 & 71.65 / 74.90 \\
(3) & \xmark & \cmark & \cmark & \textbf{7.14} / 6.70 & \textbf{77.80} / 80.24 \\
(4) & \cmark & \xmark & \xmark & 27.38 / 22.69 & 27.53 / 29.72 \\
(5) & \cmark & \xmark & \cmark & 18.54 / 17.64 & 36.32 / 39.39 \\
(6) & \cmark & \cmark & \xmark & 7.70 / 6.97 & 74.26 / 78.36 \\
(7) & \cmark & \cmark & \cmark & 7.20 / \textbf{6.59} & 77.40 / \textbf{80.90} \\
\bottomrule
\end{tabular}
\end{table}

Additionally, to quantify the individual contributions of FLOPs, parameter count, and the memory access proxy, we test different combinations of these features on FeatureFormer. Results are shown in \cref{tab:enc_makeup}. Each individual feature provides a measurable improvement over the baseline that uses none of the three. In most cases, combinations of two features outperform single feature variants, with the exception of (4), parameters plus memory, which performs slightly worse than (2), simply using the memory proxy alone. The strongest results are achieved by (7), FLOPs plus parameters, and (8), the full combination of FLOPs, parameters, and memory. Although (7) is marginally inferior to the full configuration which we utilized, their close performance indicates that FLOPs and parameters account for most of the predictive gain in $F_{comp}$, while the memory proxy provides an additional but smaller improvement.

\begin{table}[h]
\centering
\renewcommand\bfdefault{b}
\caption{\textbf{Results of different makeups of $F_{comp}$ on FeatureFormer on NNLQ \citep{liu2022nnlqp} (out-of-domain).} Memory refers to a proxy of memory operations within the model, consisting of the multiplied output shape + parameter count.}
\label{tab:enc_makeup}
\setlength{\tabcolsep}{8pt}
\begin{tabular}{c c c c c c}
\toprule
\multirow{2}{*{\#}} &
\multicolumn{3}{c}{Components} &
\multicolumn{2}{c}{Metrics (avg / best)} \\
\cmidrule(lr){2-4} \cmidrule(lr){5-6}
 & FLOPs & Parameters & Memory & MAPE $\downarrow$ & Acc (10\%) $\uparrow$ \\
\midrule
(1) & \xmark & \xmark & \xmark & 7.70 / 6.97 & 74.26 / 78.36 \\
(2) & \xmark & \xmark & \cmark & 7.46 / 7.08 & 75.89 / 78.20 \\
(3) & \xmark & \cmark & \xmark & 7.60 / 6.87 & 75.75 / 79.91 \\
(4) & \xmark & \cmark & \cmark & 7.57 / 6.96 & 75.30 / 78.92 \\
(5) & \cmark & \xmark & \xmark & 7.55 / 6.93 & 74.87 / 78.38 \\
(6) & \cmark & \xmark & \cmark & 7.47 / 7.00 & 75.67 / 78.65 \\
(7) & \cmark & \cmark & \xmark & 7.23 / 6.82 & 77.40 / 79.35 \\
(8) & \cmark & \cmark & \cmark & \textbf{7.20 / 6.59} & \textbf{77.40 / 80.90} \\
\bottomrule
\end{tabular}
\end{table}

\subsection{Model Size Comparison}
\label{sec:scaling}
While the advantages of FeatureFormer are demonstrated in \cref{sec:experiments}, it is worth noting that FeatureFormer possesses a larger parameter footprint than the next closest model, NN-Former. To verify that this superiority is attributable to architecture and is not merely a consequence of increased model size, \cref{tab:scaling} evaluates both models under parameter-matched configurations on NNLQ latency prediction. Specifically, we compare a reduced two-layer variant of FeatureFormer against the baseline NN-Former, as well as an upscaled version of NN-Former scaled to match the original FeatureFormer's size. To ensure a fair comparison and optimal performance, NN-Former is augmented with FeatureFormer's encoding, as in \cref{tab:enc_ablation_latency}. Under both equal-parameter settings, FeatureFormer consistently outperforms NN-Former; notably, even the downscaled FeatureFormer yields better performance than the upscaled NN-Former.

\begin{table}[]
\centering
\caption{Comparison of NN-Former \citep{xu2025nn} and FeatureFormer at comparable parameter counts on NNLQ \citep{liu2022nnlqp}, both using node-level static features. Symbols next to Layers indicate default configuration (\textbf{--}), scaled up ($\uparrow$), or scaled down ($\downarrow$). The scaled up/down configuration for a given model is compared against the default configuration of the other.} 

\label{tab:scaling}
\begin{tabular}{@{}lcccc@{}}
\toprule
Model & \multicolumn{1}{c}{Layers} & \multicolumn{1}{c}{Parameters (Millions)} & \multicolumn{1}{c}{MAPE} & \multicolumn{1}{c}{Acc (10\%)} \\ \midrule
\multirow{2}{*}{NN-Former \citep{xu2025nn}} & 2 \textbf{--} & 10.1 & 7.47 / 7.09 & 76.63 / \textbf{79.22} \\
 & 3 $\uparrow$ & 14.3 & 7.53 / 7.21 & 76.31 / 77.96 \\ \midrule
\multirow{2}{*}{FeatureFormer} & 2 $\downarrow$ & 8.53 & \textbf{7.45} /\textbf{ 7.03} & \textbf{76.64} / 79.14 \\
 & 4 \textbf{--} & 15.4 & \textbf{7.20} / \textbf{6.59} & \textbf{77.40} / \textbf{80.90} \\ \bottomrule
\end{tabular}
\end{table}

\section{Conclusion and Future Work}
In this work, we introduced FeatureFormer, a neural performance predictor that combines a gated graph self attention mechanism with an explicit node-level computational encoding. By incorporating FLOPs, parameter counts, and a proxy for memory operations directly into node representations, our approach addresses a limitation of prior graph based predictors, which lack information about the computational cost of individual operations. Our experiments show that this encoding improves prediction accuracy and reliability across both latency and energy prediction tasks. These gains hold not only for FeatureFormer but also when the proposed encoding is applied to several recent methods, suggesting that the approach is not specific to a single architecture. FeatureFormer performs competitively across multiple datasets and evaluation settings, including out-of-domain scenarios.

These results suggest that node-level computational encodings are a useful direction for improving neural performance prediction, and enable future work in hardware aware neural architecture search. One direction worth exploring is a foundation model approach to performance prediction: training a large, hardware-aware model that leverages GPU-specific characteristics to generalize performance predictions across different GPU architectures. The node-level static features introduced in this work could serve as a starting point for this direction.

\section*{Acknowledgments}
This work was supported by the Office of Naval Research (ONR) under contract No. N00014-23-C-1052. The support of the ONR is gratefully acknowledged. Any opinions, findings, conclusions, or recommendations expressed in this material are those of the authors and do not necessarily reflect the views of the United States Navy. This article was approved for public release with DCN \# 2026-8-12-2699 and Distribution Statement A: Approved for public release: distribution is unlimited.

\bibliography{main}
\bibliographystyle{tmlr}

\newpage
\appendix
\crefalias{section}{appendix}

\section{Experiment Details}
\label{sec:exp-details}
\subsection{Training Hyperparameters}
Across all experiments, FeatureFormer was trained using 4 layers. Each layer consisted of our Gated Graph Self Attention (GGSA) block followed by a standard ReLU feed-forward network. We set \( d_{model} = 512 \) and \( d_{ffn} = 2048 \), following prior work \citep{yi2023narv2,xu2025nn}. Unless otherwise stated in ablation studies, the input vector had a total length of 222. This included 32 dimensions for \( F_{op} \), 160 dimensions for \( F_{attr} \) (as in NN-Former \citep{xu2025nn}), and an additional 30 dimensions for \( F_{comp} \).

For training, we used a batch size of 16 and trained for 75 epochs. The first 10\% of training steps were used for learning rate warm-up \citep{goyal2017accurate}, increasing the learning rate from \(1 \times 10^{-4}\) to \(1 \times 10^{-3}\). After the warm-up phase, we applied a cosine decay schedule for the rest of training. We used the AdamW optimizer \citep{loshchilov2017decoupled} with decay rates of (0.9, 0.999). Consistent with previous work \citep{yi2023narv2,xu2025nn}, we trained the model using mean squared error loss. A dropout rate of 0.05 was used in all experiments.

When evaluating prior state-of-the-art methods, including NNLP \citep{liu2022nnlqp}, NAR-Former V2 \citep{yi2023narv2}, and NN-Former \citep{xu2025nn}, we used the training hyperparameters and official implementations released with their papers. The only exception was in experiments studying the effect of \( F_{comp} \) (\cref{tab:latency_app,tab:energy_app}). In these cases, we performed a small grid search over the sizes of \( F_{op} \) and \( F_{comp} \) to ensure each model had a fair opportunity to benefit from \( F_{comp} \) and compete with FeatureFormer. Among the compared models, only NAR-Former V2 improved with a modified input configuration. Specifically, increasing the size of \( F_{attr} \) from 120 to 160 to match NN-Former and FeatureFormer led to better performance.

\subsection{Training Cost}
In total, each experimental run of FeatureFormer required approximately 2 hours of training on a 40\,GB A100 GPU. By an experimental run, we mean training a single model with a single random seed. Therefore, evaluating a full table with 10 different models and 10 seeds each would require 100 individual runs. Based on this setup, we estimate that collecting all results reported in this paper required approximately 2{,}000 A100 GPU-hours in total.

\section{Implementation Details} 
\label{sec:imp_details} 
To clarify the implementation of our GGSA layer, we provide pseudocode in \cref{fig:pseudo}. As outlined in \cref{fig:pseudo}, the introduction of gating scores allows the model to nonlinearly combine the outputs of differently masked attention heads. This added flexibility improves accuracy, as demonstrated in \cref{tab:gating_app}.

\begin{figure}
\caption{Pseudocode of our proposed GGSA layer.}
\label{fig:pseudo}
\large
\begin{lstlisting}[]
def GGSA(X, A):
    # X: Input (B, L, D), A: Adj Matrix (B, L, L)

    Q, K, V = X @ W_q, X @ W_k, X @ W_v  # (B, L, D)
    
    Q, K, V = reshape((Q, K, V), (B, 4, L, D // 4))  # (B, 4, L, D//4)

    gating_scores = sigmoid(X @ W_G)  # (B, L, D)

    scores = (Q @ K.T) / sqrt(D // 4)  # (B, 4, L, L)

    masks = concat(A, A.T, A @ A.T, A.T @ A)  # (B, 4, L, L)
    
    masks = masks + Identity(L, L)                  
    masks = Binarize(masks)                         

    scores = scores * masks                       
    scores = softmax(scores)                      

    out = scores @ V  # (B, 4, L, D//4)
    
    out = reshape(out, (B, L, D))  # (B, L, D)

    out = out * gating_scores                     

    out = out @ W_o   

    return out
\end{lstlisting}
\end{figure}

\section{NNEQ Statistics}
\begin{figure}[]
\centering
\includegraphics[width=1\linewidth]{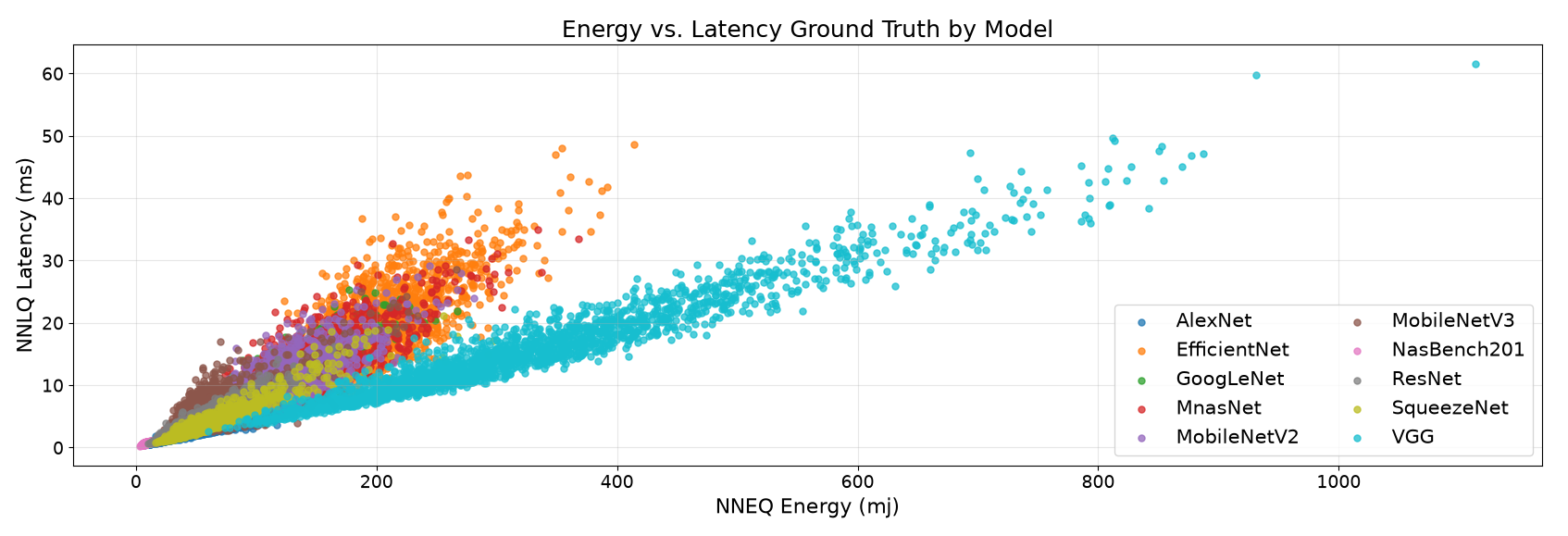}
\caption{A comparison of NNLQ \citep{liu2022nnlqp} ground truth latency and NNEQ ground truth energy across individual models. The relationship between these metrics is neither fully linear nor fully proportional, emphasizing the distinct nature of the two predictive tasks.}
\label{fig:energy-vs-latency}
\end{figure}

To provide further insight into the differences between NNLQ \citep{liu2022nnlqp} and NNEQ, further analysis is provided in \cref{fig:energy-vs-latency} and \cref{tab:dataset-stats}. \Cref{fig:energy-vs-latency} provides a plot of NNLQ latency measurements \textit{v.s.} NNEQ energy measurements, detailing how the two are correlated, while \cref{tab:dataset-stats} provides detailed statistics on a per-family and per-dataset basis for NNEQ and NNLQ.

\begin{table*}
\small
\renewcommand\bfdefault{b}
\centering
\caption{\textbf{Summary of NNLQ \citep{liu2022nnlqp} and NNEQ dataset statistics by model family.} Statistics are computed over the ground truth values of all samples in each family. "All" aggregates across all families within a dataset.}
\label{tab:dataset-stats}
{
\setlength{\tabcolsep}{4pt}
\begin{tabular}{@{}c|l|cccccc@{}}
\toprule
Dataset & Architecture & Mean & Median & Min & Max & Std & Count \\
\midrule

\multicolumn{1}{c|}{\multirow{11}{*}{\rotatebox{90}{NNEQ}}}
& AlexNet        & 51.45  & 47.56  & 6.64  & 140.04  & 24.43  & 2000  \\
& EfficientNet   & 189.44 & 184.55 & 33.70 & 414.06  & 45.44  & 2000  \\
& GoogLeNet      & 111.69 & 108.26 & 27.60 & 267.25  & 29.86  & 2000  \\
& MnasNet        & 128.72 & 121.62 & 17.67 & 367.98  & 42.30  & 2000  \\
& MobileNetV2    & 113.31 & 108.84 & 16.09 & 281.15  & 33.17  & 2000  \\
& MobileNetV3    & 52.14  & 46.46  & 9.94  & 266.29  & 25.13  & 2000  \\
& NasBench201    & 8.28   & 8.18   & 3.07  & 12.98   & 1.53   & 2000  \\
& ResNet         & 47.14  & 42.69  & 10.41 & 192.91  & 21.90  & 2000  \\
& SqueezeNet     & 56.83  & 49.73  & 16.46 & 297.20  & 29.54  & 2000  \\
& VGG            & 323.23 & 295.34 & 60.27 & 1113.80 & 143.14 & 2000  \\
\cmidrule{2-8}
& All            & 108.22 & 81.56  & 3.07  & 1113.80 & 102.43 & 20000 \\

\midrule

\multicolumn{1}{c|}{\multirow{11}{*}{\rotatebox{90}{NNLQ}}}
& AlexNet        & 2.63  & 2.45  & 0.41 & 10.39 & 1.27 & 2000  \\
& EfficientNet   & 17.73 & 16.68 & 2.29 & 48.61 & 6.73 & 2000  \\
& GoogLeNet      & 9.85  & 9.38  & 1.35 & 25.35 & 3.64 & 2000  \\
& MnasNet        & 10.63 & 9.49  & 1.14 & 34.97 & 5.21 & 2000  \\
& MobileNetV2    & 9.45  & 8.63  & 1.08 & 29.21 & 4.09 & 2000  \\
& MobileNetV3    & 5.02  & 4.24  & 0.69 & 28.61 & 3.02 & 2000  \\
& NasBench201    & 0.70  & 0.70  & 0.23 & 1.10  & 0.12 & 2000  \\
& ResNet         & 3.67  & 3.16  & 0.73 & 18.23 & 2.12 & 2000  \\
& SqueezeNet     & 3.57  & 2.86  & 0.87 & 21.16 & 2.48 & 2000  \\
& VGG            & 15.61 & 13.93 & 2.53 & 61.53 & 7.99 & 2000  \\
\cmidrule{2-8}
& All            & 7.89  & 5.85  & 0.23 & 61.53 & 6.94 & 20000 \\

\bottomrule
\end{tabular}
}

\end{table*}

\section{Expanded Results}
Due to space constraints, several complete result tables could not be included in the main paper and are therefore provided in this section. \Cref{tab:latency_app,tab:energy_app} present the full results of the $F_{comp}$ ablation studies, showing that node-level computational information benefits a range of state-of-the-art architectures. \Cref{tab:gating_app} reports the impact of gating within our proposed GGSA layer, demonstrating its effectiveness. \Cref{tab:app_mask} highlights the importance of masking each attention head separately, compared to using a shared or global mask. Finally, \cref{tab:app_enc,tab:app_comp} provide detailed ablation studies on the design of our encoding methodology.

\begin{table*}
\renewcommand\bfdefault{b}
\scriptsize
\centering
\caption{\textbf{Full performance of various models on NNLQ \citep{liu2022nnlqp} (out-of-domain) with node-level computational information ($F_{comp}$) added.} Results for each model are reported over 10 independent trials.}
\label{tab:latency_app}
\begin{tabular}{@{}cllllclclclcl@{}}
\toprule
Metric &  & Test Domain &  &  & \begin{tabular}[c]{@{}c@{}}NNLP \\\cite{liu2022nnlqp}\\ (avg / best)\end{tabular} &  & \begin{tabular}[c]{@{}c@{}}NAR-Former V2 \\\cite{yi2023narv2}\\ (avg / best)\end{tabular} &  & \begin{tabular}[c]{@{}c@{}}NN-Former \\\cite{xu2025nn}\\ (avg / best)\end{tabular} &  & \begin{tabular}[c]{@{}c@{}}FeatureFormer\\ (avg / best)\end{tabular} &  \\ \midrule
\multicolumn{1}{c|}{\multirow{22}{*}{\rotatebox{90}{MAPE $\downarrow$}}} &  & AlexNet & \multicolumn{1}{l|}{} &  & 10.93 / 9.99 &  & 13.16 / 11.74 &  & 10.39 / 9.84 &  & \textbf{10.15 / 9.51} &  \\
\multicolumn{1}{c|}{} &  &              & \multicolumn{1}{l|}{} &  &({\color[HTML]{C82F09}+0.29} / {\color[HTML]{C82F09}+0.28})&  &({\color[HTML]{349C07}-11.12} / {\color[HTML]{349C07}-6.55})&  &({\color[HTML]{349C07}-1.08} / {\color[HTML]{349C07}-1.33})&  &({\color[HTML]{349C07}-1.68} / {\color[HTML]{349C07}-1.84})&  \\[2pt]
\multicolumn{1}{c|}{} &  & EfficientNet & \multicolumn{1}{l|}{} &  & 19.90 / 15.15 &  &  6.13 / 5.41   &  &  \textbf{4.15} / 3.87   &  & 4.61 / \textbf{3.53} &  \\
\multicolumn{1}{c|}{} &  &              & \multicolumn{1}{l|}{} &  &({\color[HTML]{349C07}-1.56} / {\color[HTML]{349C07}-3.57})&  &({\color[HTML]{349C07}-7.07} / {\color[HTML]{349C07}-5.96})&  &({\color[HTML]{349C07}-0.98} / {\color[HTML]{349C07}-0.94})&  &({\color[HTML]{349C07}-0.05} / {\color[HTML]{349C07}-0.27})&  \\[2pt]
\multicolumn{1}{c|}{} &  & GoogLeNet & \multicolumn{1}{l|}{}    &  & 11.26 / 10.00 &  &  12.59 / 8.48  &  &  5.62 / 5.08   &  & \textbf{4.85 / 4.77} &  \\
\multicolumn{1}{c|}{} &  &              & \multicolumn{1}{l|}{} &  &({\color[HTML]{349C07}-2.02} / {\color[HTML]{349C07}-0.90})&  &({\color[HTML]{C82F09}+5.98} / {\color[HTML]{C82F09}+2.33})&  &({\color[HTML]{349C07}-1.12} / {\color[HTML]{349C07}-1.57})&  &({\color[HTML]{349C07}-0.16} / {\color[HTML]{349C07}-0.01})&  \\[2pt]
\multicolumn{1}{c|}{} &  & MnasNet & \multicolumn{1}{l|}{}      &  & 5.47 / 5.08   &  &  2.62 / 2.46   &  &  2.25 / 2.17   &  & \textbf{1.91 / 1.79} &  \\
\multicolumn{1}{c|}{} &  &              & \multicolumn{1}{l|}{} &  &({\color[HTML]{349C07}-6.60} / {\color[HTML]{349C07}-5.78})&  &({\color[HTML]{349C07}-4.54} / {\color[HTML]{349C07}-3.47})&  &({\color[HTML]{349C07}-0.46} / {\color[HTML]{349C07}-0.37})&  &({\color[HTML]{349C07}-0.20} / {\color[HTML]{349C07}-0.10})&  \\[2pt]
\multicolumn{1}{c|}{} &  & MobileNetV2 & \multicolumn{1}{l|}{}  &  & 8.01 / 5.73   &  &  4.72 / 4.56   &  &  4.69 / 4.33   &  & \textbf{2.90 / 2.30} &  \\
\multicolumn{1}{c|}{} &  &              & \multicolumn{1}{l|}{} &  &({\color[HTML]{349C07}-0.86} / {\color[HTML]{349C07}-1.61})&  &({\color[HTML]{349C07}-2.01} / {\color[HTML]{349C07}-1.09})&  &({\color[HTML]{C82F09}+0.52} / {\color[HTML]{C82F09}+0.67})&  &({\color[HTML]{349C07}-0.53} / {\color[HTML]{349C07}-0.37})&  \\[2pt]
\multicolumn{1}{c|}{} &  & MobileNetV3 & \multicolumn{1}{l|}{}  &  & 10.38 / 9.31  &  &  8.49 / 8.12   &  &  8.16 / 7.84   &  & \textbf{8.01 / 7.71 }&  \\
\multicolumn{1}{c|}{} &  &              & \multicolumn{1}{l|}{} &  &({\color[HTML]{349C07}-4.19} / {\color[HTML]{349C07}-3.86})&  &({\color[HTML]{349C07}-0.57} / {\color[HTML]{349C07}-0.60})&  &({\color[HTML]{349C07}-0.91} / {\color[HTML]{349C07}-1.19})&  &({\color[HTML]{349C07}-0.58} / {\color[HTML]{349C07}-0.26})&  \\[2pt]
\multicolumn{1}{c|}{} &  & NasBench201 & \multicolumn{1}{l|}{}  &  & 13.88 / 6.02  &  &  7.52 / 5.64   &  &  7.83 / 5.37   &  & \textbf{6.79 / 4.98} &  \\
\multicolumn{1}{c|}{} &  &              & \multicolumn{1}{l|}{} &  &({\color[HTML]{C82F09}+4.28} / {\color[HTML]{349C07}-2.17})&  &({\color[HTML]{349C07}-1.69} / {\color[HTML]{349C07}-2.25})&  &({\color[HTML]{349C07}0.00} / {\color[HTML]{349C07}-2.34})&  &({\color[HTML]{349C07}-0.28} / {\color[HTML]{C82F09}+0.96})&  \\[2pt]
\multicolumn{1}{c|}{} &  & ResNet & \multicolumn{1}{l|}{}       &  & 6.68 / 6.44   &  &  \textbf{6.07 / 5.62}   &  &  6.77 / 6.46   &  & 7.98 / 7.10 &  \\
\multicolumn{1}{c|}{} &  &              & \multicolumn{1}{l|}{} &  &({\color[HTML]{349C07}-0.86} / {\color[HTML]{349C07}-0.68})&  &({\color[HTML]{349C07}-0.73} / {\color[HTML]{349C07}-0.82})&  &({\color[HTML]{349C07}-0.72} / {\color[HTML]{349C07}-0.92})&  &({\color[HTML]{349C07}-0.98} / {\color[HTML]{349C07}-0.62})&  \\[2pt]
\multicolumn{1}{c|}{} &  & SqueezeNet & \multicolumn{1}{l|}{}   &  & 8.42 / 7.66   &  &  7.69 / 6.49   &  &  \textbf{6.38 / 5.42}   &  & 6.41 / 5.86 &  \\
\multicolumn{1}{c|}{} &  &              & \multicolumn{1}{l|}{} &  &({\color[HTML]{349C07}-1.42} / {\color[HTML]{349C07}-1.86})&  &({\color[HTML]{C82F09}+0.61} / {\color[HTML]{349C07}-0.07})&  &({\color[HTML]{349C07}-2.70} / {\color[HTML]{349C07}-1.63})&  &({\color[HTML]{349C07}-0.59} / {\color[HTML]{349C07}-0.17})&  \\[2pt]
\multicolumn{1}{c|}{} &  & VGG & \multicolumn{1}{l|}{}          &  & \textbf{5.74 / 5.67}   &  &  20.42 / 19.42 &  &  19.70 / 19.03 &  & 18.79 / 17.30 &  \\
\multicolumn{1}{c|}{} &  &              & \multicolumn{1}{l|}{} &  &({\color[HTML]{349C07}-1.86} / {\color[HTML]{349C07}-1.51})&  &({\color[HTML]{C82F09}+5.02} / {\color[HTML]{C82F09}+5.16})&  &({\color[HTML]{349C07}-0.42} / {\color[HTML]{349C07}-0.61})&  &({\color[HTML]{C82F09}+1.14} / {\color[HTML]{C82F09}0.42})&  \\\cmidrule{2-13}
\multicolumn{1}{c|}{} &  & Average & \multicolumn{1}{l|}{}      &  & 10.07 / 8.11  &  &  8.94 / 7.79   &  &  7.60 / 6.94   &  & \textbf{7.24 / 6.48} &  \\
\multicolumn{1}{c|}{} &  &              & \multicolumn{1}{l|}{} &  &({\color[HTML]{349C07}-1.48} / {\color[HTML]{349C07}-2.16})&  &({\color[HTML]{349C07}-1.61} / {\color[HTML]{349C07}-1.33})&  &({\color[HTML]{349C07}-0.79} / {\color[HTML]{349C07}-1.02})&  &({\color[HTML]{349C07}-0.39} / {\color[HTML]{349C07}-0.23})&  \\\midrule

\multicolumn{1}{c|}{\multirow{22}{*}{\rotatebox{90}{Acc (10\%) $\uparrow$}}} &  & AlexNet & \multicolumn{1}{l|}{} &  & 55.72 / 59.65 &  & 50.36	 / 55.20 &  & 60.24	 / 64.00 &  & \textbf{60.76	/ 64.40} &  \\
\multicolumn{1}{c|}{} &  &              & \multicolumn{1}{l|}{} &  &({\color[HTML]{C82F09}-3.35} / {\color[HTML]{C82F09}-4.75})&  &({\color[HTML]{349C07}+25.71} / {\color[HTML]{349C07}+26.60})&  &({\color[HTML]{349C07}+4.15} / {\color[HTML]{349C07}+6.90})&  &({\color[HTML]{349C07}+5.23} / {\color[HTML]{349C07}+6.90})&  \\[2pt]
\multicolumn{1}{c|}{} &  & EfficientNet & \multicolumn{1}{l|}{} &  & 26.55 / 42.15 &  &  80.70		/ 87.10   &  &  \textbf{93.36}	/ 94.80   &  & 90.59	/ 	\textbf{95.95} &  \\
\multicolumn{1}{c|}{} &  &              & \multicolumn{1}{l|}{} &  &({\color[HTML]{349C07}+1.18} / {\color[HTML]{349C07}+13.35})&  &({\color[HTML]{349C07}+36.69} / {\color[HTML]{349C07}+36.90})&  &({\color[HTML]{349C07}+2.51} / {\color[HTML]{349C07}+3.90})&  &({\color[HTML]{C82F09}-0.31} / {\color[HTML]{349C07}+0.65})&  \\[2pt]
\multicolumn{1}{c|}{} &  & GoogLeNet & \multicolumn{1}{l|}{}    &  & 44.34		/ 53.35 &  &  46.75		/ 66.75  &  &  85.39		/ 88.00   &  & \textbf{89.71		/ 90.55} &  \\
\multicolumn{1}{c|}{} &  &              & \multicolumn{1}{l|}{} &  &({\color[HTML]{349C07}+8.04} / {\color[HTML]{349C07}+4.60})&  &({\color[HTML]{C82F09}-33.36} / {\color[HTML]{C82F09}-16.60})&  &({\color[HTML]{349C07}+4.96} / {\color[HTML]{349C07}+4.60})&  &({\color[HTML]{349C07}+1.26} / {\color[HTML]{349C07}+0.70})&  \\[2pt]
\multicolumn{1}{c|}{} &  & MnasNet & \multicolumn{1}{l|}{}      &  & 85.16		/ 88.25   &  &  98.89		/ 99.25   &  &  99.40		/ 99.55   &  & \textbf{99.76		/ 99.85} &  \\
\multicolumn{1}{c|}{} &  &              & \multicolumn{1}{l|}{} &  &({\color[HTML]{349C07}+29.27} / {\color[HTML]{349C07}+27.00})&  &({\color[HTML]{349C07}+25.43} / {\color[HTML]{349C07}+17.65})&  &({\color[HTML]{349C07}+0.75} / {\color[HTML]{349C07}+0.85})&  &({\color[HTML]{349C07}+0.51} / {\color[HTML]{349C07}+0.20})&  \\[2pt]
\multicolumn{1}{c|}{} &  & MobileNetV2 & \multicolumn{1}{l|}{}  &  & 68.92		/ 83.45   &  &  87.02		/ 88.80  &  &  88.11		/ 91.00   &  & \textbf{97.31		/ 99.65} &  \\
\multicolumn{1}{c|}{} &  &              & \multicolumn{1}{l|}{} &  &({\color[HTML]{349C07}+5.89} / {\color[HTML]{349C07}+10.95})&  &({\color[HTML]{349C07}+8.57} / {\color[HTML]{349C07}+5.00})&  &({\color[HTML]{C82F09}-6.79} / {\color[HTML]{C82F09}-5.85})&  &({\color[HTML]{349C07}+1.58} / {\color[HTML]{349C07}+0.15})&  \\[2pt]
\multicolumn{1}{c|}{} &  & MobileNetV3 & \multicolumn{1}{l|}{}  &  & 59.09		/ 65.40  &  &  69.75		/ 72.05   &  &  72.16		/ 74.70   &  & \textbf{73.38		/ 75.75} &  \\
\multicolumn{1}{c|}{} &  &              & \multicolumn{1}{l|}{} &  &({\color[HTML]{349C07}+15.83} / {\color[HTML]{349C07}+15.75})&  &({\color[HTML]{349C07}+1.31} / {\color[HTML]{349C07}+1.55})&  &({\color[HTML]{C82F09}-2.03} / {\color[HTML]{349C07}+0.40})&  &({\color[HTML]{349C07}+3.09} / {\color[HTML]{349C07}+1.45})&  \\[2pt]
\multicolumn{1}{c|}{} &  & NasBench201 & \multicolumn{1}{l|}{}  &  & 58.17		/ 83.10  &  &  73.07		/ 84.55  &  &  70.39		/ 86.95   &  & \textbf{77.28		/ 90.25} &  \\
\multicolumn{1}{c|}{} &  &              & \multicolumn{1}{l|}{} &  &({\color[HTML]{C82F09}-2.53} / {\color[HTML]{349C07}+12.50})&  &({\color[HTML]{349C07}+9.94} / {\color[HTML]{349C07}+12.85})&  &({\color[HTML]{349C07}+0.48} / {\color[HTML]{349C07}+15.85})&  &({\color[HTML]{349C07}+1.84} / {\color[HTML]{C82F09}-5.15})&  \\[2pt]
\multicolumn{1}{c|}{} &  & ResNet & \multicolumn{1}{l|}{}       &  & 77.35		/ 78.95   &  &  \textbf{81.63		/ 84.35}   &  &  76.47		/ 79.15   &  & 69.01		/ 75.25 &  \\
\multicolumn{1}{c|}{} &  &              & \multicolumn{1}{l|}{} &  &({\color[HTML]{349C07}+4.47} / {\color[HTML]{349C07}+2.55})&  &({\color[HTML]{349C07}+4.38} / {\color[HTML]{349C07}+4.65})&  &({\color[HTML]{349C07}+5.64} / {\color[HTML]{349C07}+7.60})&  &({\color[HTML]{349C07}+6.60} / {\color[HTML]{349C07}+6.55})&  \\[2pt]
\multicolumn{1}{c|}{} &  & SqueezeNet & \multicolumn{1}{l|}{}   &  & 66.74		/ 71.00  &  &  71.34		/ 79.05   &  &  \textbf{79.57		/ 86.35}   &  & 79.05		/ 83.55 &  \\
\multicolumn{1}{c|}{} &  &              & \multicolumn{1}{l|}{} &  &({\color[HTML]{349C07}+8.05} / {\color[HTML]{349C07}+10.60})&  &({\color[HTML]{C82F09}-3.68} / {\color[HTML]{C82F09}-0.20})&  &({\color[HTML]{349C07}+1.72} / {\color[HTML]{349C07}+5.40})&  &({\color[HTML]{349C07}+4.33} / {\color[HTML]{349C07}+0.75})&  \\[2pt]
\multicolumn{1}{c|}{} &  & VGG & \multicolumn{1}{l|}{}          &  & \textbf{84.53		/ 85.20}   &  &  34.88		/ 37.45 &  &  31.92		/ 33.85 &  & 35.04		/ 39.25 &  \\
\multicolumn{1}{c|}{} &  &              & \multicolumn{1}{l|}{} &  &({\color[HTML]{349C07}+13.49} / {\color[HTML]{349C07}+11.45})&  &({\color[HTML]{C82F09}-10.34} / {\color[HTML]{C82F09}-7.85})&  &({\color[HTML]{349C07}+2.52} / {\color[HTML]{349C07}+4.00})&  &({\color[HTML]{349C07}+1.31} / {\color[HTML]{349C07}+2.60})&  \\\cmidrule{2-13}
\multicolumn{1}{c|}{} &  & Average & \multicolumn{1}{l|}{}      &  & 62.66		/ 71.05  &  &  69.43		/ 75.46  &  &  75.70		/ 79.84   &  & \textbf{77.19		/ 81.45} &  \\
\multicolumn{1}{c|}{} &  &              & \multicolumn{1}{l|}{} &  &({\color[HTML]{349C07}+8.03} / {\color[HTML]{349C07}+10.40})&  &({\color[HTML]{349C07}+6.46} / {\color[HTML]{349C07}+8.06})&  &({\color[HTML]{349C07}+1.39} / {\color[HTML]{349C07}+4.37})&  &({\color[HTML]{349C07}+2.54} / {\color[HTML]{349C07}+1.48})&  \\ \bottomrule
\end{tabular}
\end{table*}

\begin{table*}
\renewcommand\bfdefault{b}
\scriptsize
\centering
\caption{\textbf{Full performance of various models on NNEQ (out-of-domain) with node-level computational information ($F_{comp}$) added.} Results for each model are reported over 10 independent trials.}
\label{tab:energy_app}
\begin{tabular}{@{}cllllclclclcl@{}}
\toprule
Metric &  & Test Domain &  &  & \begin{tabular}[c]{@{}c@{}}NNLP \\\cite{liu2022nnlqp}\\ (avg / best)\end{tabular} &  & \begin{tabular}[c]{@{}c@{}}NAR-Former V2 \\\cite{yi2023narv2}\\ (avg / best)\end{tabular} &  & \begin{tabular}[c]{@{}c@{}}NN-Former \\\cite{xu2025nn}\\ (avg / best)\end{tabular} &  & \begin{tabular}[c]{@{}c@{}}FeatureFormer\\ (avg / best)\end{tabular} &  \\ \midrule
\multicolumn{1}{c|}{\multirow{22}{*}{\rotatebox{90}{MAPE $\downarrow$}}} &  & AlexNet & \multicolumn{1}{l|}{} &  & 11.47 / 8.43 &  & 7.96 / 5.42 &  & 8.17 / 5.93 &  & \textbf{7.29 / 5.42} &  \\
\multicolumn{1}{c|}{} &  &              & \multicolumn{1}{l|}{} &  &({\color[HTML]{C82F09}+0.77} / {\color[HTML]{C82F09}+0.37})&  &({\color[HTML]{349C07}-8.73} / {\color[HTML]{349C07}-7.02})&  &({\color[HTML]{C82F09}+0.43} / {\color[HTML]{349C07}-0.01})&  &({\color[HTML]{C82F09}+0.22} / {\color[HTML]{349C07}-0.72})&  \\[2pt]
\multicolumn{1}{c|}{} &  & EfficientNet & \multicolumn{1}{l|}{} &  & 5.42 / 4.86 &  &  \textbf{2.71 / 2.47}   &  &  3.02 / 2.79   &  & 2.89 / 2.47 &  \\
\multicolumn{1}{c|}{} &  &              & \multicolumn{1}{l|}{} &  &({\color[HTML]{349C07}-1.03} / {\color[HTML]{349C07}-1.16})&  &({\color[HTML]{349C07}-1.78} / {\color[HTML]{349C07}-1.62})&  &({\color[HTML]{349C07}-0.22} / {\color[HTML]{349C07}-0.01})&  &({\color[HTML]{349C07}-1.07} / {\color[HTML]{349C07}-0.67})&  \\[2pt]
\multicolumn{1}{c|}{} &  & GoogLeNet & \multicolumn{1}{l|}{}    &  & 3.67 / 3.42 &  &  3.83 / 3.53  &  &  \textbf{3.60} / \textbf{3.25}   &  & 3.63 / 3.35 &  \\
\multicolumn{1}{c|}{} &  &              & \multicolumn{1}{l|}{} &  &({\color[HTML]{349C07}-0.23} / {\color[HTML]{349C07}-0.07})&  &({\color[HTML]{C82F09}+0.39} / {\color[HTML]{C82F09}+0.22})&  &({\color[HTML]{349C07}-0.24} / {\color[HTML]{349C07}-0.20})&  &({\color[HTML]{C82F09}+0.03} / {\color[HTML]{C82F09}+0.02})&  \\[2pt]
\multicolumn{1}{c|}{} &  & MnasNet & \multicolumn{1}{l|}{}      &  & 9.03 / 6.45   &  &  3.41 / 3.07   &  &  3.33 / 2.79   &  & \textbf{3.10 / 2.72} &  \\
\multicolumn{1}{c|}{} &  &              & \multicolumn{1}{l|}{} &  &({\color[HTML]{349C07}-0.15} / {\color[HTML]{349C07}-1.36})&  &({\color[HTML]{349C07}-1.08} / {\color[HTML]{349C07}-0.94})&  &({\color[HTML]{C82F09}+0.04} / {\color[HTML]{C82F09}+0.08})&  &({\color[HTML]{C82F09}+0.08} / {\color[HTML]{C82F09}+0.04})&  \\[2pt]
\multicolumn{1}{c|}{} &  & MobileNetV2 & \multicolumn{1}{l|}{}  &  & 4.79 / 3.94   &  &  3.06 / 2.91   &  &  3.20 / 2.93   &  & \textbf{3.04 / 2.70} &  \\
\multicolumn{1}{c|}{} &  &              & \multicolumn{1}{l|}{} &  &({\color[HTML]{349C07}-1.19} / {\color[HTML]{349C07}-1.01})&  &({\color[HTML]{349C07}-0.71} / {\color[HTML]{349C07}-0.51})&  &({\color[HTML]{349C07}-0.14} / {\color[HTML]{349C07}-0.01})&  &({\color[HTML]{349C07}-0.13} / {\color[HTML]{349C07}-0.19})&  \\[2pt]
\multicolumn{1}{c|}{} &  & MobileNetV3 & \multicolumn{1}{l|}{}  &  & 12.77 / 8.93  &  &  6.18 / 5.77   &  &  \textbf{4.68 / 4.47}   &  & 5.18 / 4.79 &  \\
\multicolumn{1}{c|}{} &  &              & \multicolumn{1}{l|}{} &  &({\color[HTML]{349C07}-0.98} / {\color[HTML]{C82F09}+1.57})&  &({\color[HTML]{349C07}-2.05} / {\color[HTML]{349C07}-1.49})&  &({\color[HTML]{C82F09}+0.08} / {\color[HTML]{C82F09}+0.11})&  &({\color[HTML]{C82F09}+0.12} / {\color[HTML]{349C07}-0.07})&  \\[2pt]
\multicolumn{1}{c|}{} &  & NasBench201 & \multicolumn{1}{l|}{}  &  & 9.87 / 8.19  &  &  10.11 / 6.70   &  &  16.77 / \textbf{5.70}   &  & \textbf{8.94} / 7.45 &  \\
\multicolumn{1}{c|}{} &  &              & \multicolumn{1}{l|}{} &  &({\color[HTML]{349C07}-0.93} / {\color[HTML]{349C07}-0.17})&  &({\color[HTML]{349C07}-1.11} / {\color[HTML]{349C07}-2.39})&  &({\color[HTML]{C82F09}+1.10} / {\color[HTML]{349C07}-0.46})&  &({\color[HTML]{C82F09}0.23} / {\color[HTML]{C82F09}+0.23})&  \\[2pt]
\multicolumn{1}{c|}{} &  & ResNet & \multicolumn{1}{l|}{}       &  & 5.52 / 5.35   &  &  4.81 / 4.65   &  &  4.90 / 4.65   &  & \textbf{4.65 / 4.36} &  \\
\multicolumn{1}{c|}{} &  &              & \multicolumn{1}{l|}{} &  &({\color[HTML]{349C07}-0.17} / {\color[HTML]{349C07}-0.10})&  &({\color[HTML]{349C07}-0.50} / {\color[HTML]{349C07}-0.38})&  &({\color[HTML]{349C07}-0.10} / {\color[HTML]{C82F09}+0.03})&  &({\color[HTML]{349C07}-0.30} / {\color[HTML]{349C07}-0.34})&  \\[2pt]
\multicolumn{1}{c|}{} &  & SqueezeNet & \multicolumn{1}{l|}{}   &  & 6.71 / 6.29   &  &  4.71 / 4.47   &  &  4.86 / 4.39   &  & \textbf{4.57 / 4.15} &  \\
\multicolumn{1}{c|}{} &  &              & \multicolumn{1}{l|}{} &  &({\color[HTML]{349C07}-0.25} / {\color[HTML]{349C07}-0.05})&  &({\color[HTML]{349C07}-1.39} / {\color[HTML]{349C07}-1.01})&  &({\color[HTML]{349C07}-0.34} / {\color[HTML]{349C07}-0.36})&  &({\color[HTML]{349C07}-0.16} / {\color[HTML]{349C07}-0.30})&  \\[2pt]
\multicolumn{1}{c|}{} &  & VGG & \multicolumn{1}{l|}{}          &  & \textbf{5.59 / 4.65}   &  &  22.05 / 19.82 &  &  26.29 / 24.79 &  & 23.05 / 21.17 &  \\
\multicolumn{1}{c|}{} &  &              & \multicolumn{1}{l|}{} &  &({\color[HTML]{C82F09}+0.88} / {\color[HTML]{C82F09}+0.12})&  &({\color[HTML]{C82F09}+6.06} / {\color[HTML]{C82F09}+4.89})&  &({\color[HTML]{349C07}-0.18} / {\color[HTML]{349C07}-0.70})&  &({\color[HTML]{349C07}-0.71} / {\color[HTML]{349C07}-0.56})&  \\\cmidrule{2-13}
\multicolumn{1}{c|}{} &  & Average & \multicolumn{1}{l|}{}      &  & 7.48 / 6.05  &  &  6.88 / 5.88   &  &  7.88 / 6.17   &  & \textbf{6.63 / 5.86} &  \\
\multicolumn{1}{c|}{} &  &              & \multicolumn{1}{l|}{} &  &({\color[HTML]{349C07}-0.33} / {\color[HTML]{349C07}-0.18})&  &({\color[HTML]{349C07}-1.09} / {\color[HTML]{349C07}-1.03})&  &({\color[HTML]{C82F09}+0.04} / {\color[HTML]{349C07}-0.15})&  &({\color[HTML]{349C07}-0.17} / {\color[HTML]{349C07}-0.20})&  \\\midrule

\multicolumn{1}{c|}{\multirow{22}{*}{\rotatebox{90}{Acc (10\%) $\uparrow$}}} &  & AlexNet & \multicolumn{1}{l|}{} &  & 57.68 / 71.50 &  & 69.61	 / \textbf{87.25} &  & 66.82	 / 80.80 &  & \textbf{73.53}	/ 85.80 &  \\
\multicolumn{1}{c|}{} &  &              & \multicolumn{1}{l|}{} &  &({\color[HTML]{C82F09}-3.17} / {\color[HTML]{C82F09}-0.75})&  &({\color[HTML]{349C07}+25.88} / {\color[HTML]{349C07}+23.85})&  &({\color[HTML]{C82F09}-3.13} / {\color[HTML]{349C07}+0.75})&  &({\color[HTML]{C82F09}-0.17} / {\color[HTML]{349C07}+6.65})&  \\[2pt]
\multicolumn{1}{c|}{} &  & EfficientNet & \multicolumn{1}{l|}{} &  & 86.02 / 90.30 &  &  \textbf{99.53 / 99.95}   &  &  98.57	/ 99.65   &  & 98.62	/ 	99.75 &  \\
\multicolumn{1}{c|}{} &  &              & \multicolumn{1}{l|}{} &  &({\color[HTML]{349C07}+7.54} / {\color[HTML]{349C07}+7.90})&  &({\color[HTML]{349C07}+7.42} / {\color[HTML]{349C07}+5.20})&  &({\color[HTML]{349C07}+0.67} / {\color[HTML]{349C07}+0.55})&  &({\color[HTML]{349C07}+3.43} / {\color[HTML]{349C07}+1.35})&  \\[2pt]
\multicolumn{1}{c|}{} &  & GoogLeNet & \multicolumn{1}{l|}{}    &  & \textbf{96.65	}	/ 97.90 &  &  96.50	/\textbf{ 98.10}  &  &  96.52	/ 98.00   &  & 96.62	/ 97.95 &  \\
\multicolumn{1}{c|}{} &  &              & \multicolumn{1}{l|}{} &  &({\color[HTML]{349C07}+0.62} / {\color[HTML]{349C07}+0.35})&  &({\color[HTML]{C82F09}-0.82} / {\color[HTML]{349C07}+0.15})&  &({\color[HTML]{349C07}+1.12} / {\color[HTML]{349C07}+0.65})&  &({\color[HTML]{C82F09}-0.39} / {\color[HTML]{C82F09}-0.45})&  \\[2pt]
\multicolumn{1}{c|}{} &  & MnasNet & \multicolumn{1}{l|}{}      &  & 59.75		/ 78.40   &  &  97.88	/ 98.70   &  &  98.13 / 99.30   &  & \textbf{98.76	/ 99.75} &  \\
\multicolumn{1}{c|}{} &  &              & \multicolumn{1}{l|}{} &  &({\color[HTML]{349C07}+0.76} / {\color[HTML]{349C07}+8.45})&  &({\color[HTML]{349C07}+5.51} / {\color[HTML]{349C07}+3.30})&  &({\color[HTML]{349C07}+0.09} / {\color[HTML]{349C07}+0.10})&  &({\color[HTML]{C82F09}-0.38} / {\color[HTML]{349C07}+0.10})&  \\[2pt]
\multicolumn{1}{c|}{} &  & MobileNetV2 & \multicolumn{1}{l|}{}  &  & 89.83		/ 94.90   &  &  98.41	/ 99.00  &  &  98.16	/ 99.00   &  & \textbf{98.48		/ 99.15} &  \\
\multicolumn{1}{c|}{} &  &              & \multicolumn{1}{l|}{} &  &({\color[HTML]{349C07}+8.11} / {\color[HTML]{349C07}+6.75})&  &({\color[HTML]{349C07}+2.48} / {\color[HTML]{349C07}+1.60})&  &({\color[HTML]{349C07}+0.55} / {\color[HTML]{C82F09}+0.25})&  &({\color[HTML]{349C07}+0.13} / {\color[HTML]{349C07}+0.20})&  \\[2pt]
\multicolumn{1}{c|}{} &  & MobileNetV3 & \multicolumn{1}{l|}{}  &  & 40.80		/ 64.45  &  &  81.24	/ 84.20   &  &  \textbf{90.24	/ 91.45}   &  & 87.67		/ 89.70 &  \\
\multicolumn{1}{c|}{} &  &              & \multicolumn{1}{l|}{} &  &({\color[HTML]{349C07}+4.70} / {\color[HTML]{C82F09}-8.15})&  &({\color[HTML]{349C07}+12.80} / {\color[HTML]{349C07}+9.85})&  &({\color[HTML]{C82F09}-0.92} / {\color[HTML]{C82F09}-0.85})&  &({\color[HTML]{C82F09}-0.80} / {\color[HTML]{C82F09}-0.20})&  \\[2pt]
\multicolumn{1}{c|}{} &  & NasBench201 & \multicolumn{1}{l|}{}  &  & 60.42		/ 69.50  &  &  60.38	/ 76.90 &  &  40.34 / \textbf{84.65}   &  & \textbf{64.61	}/ 73.90 &  \\
\multicolumn{1}{c|}{} &  &              & \multicolumn{1}{l|}{} &  &({\color[HTML]{349C07}+5.26} / {\color[HTML]{349C07}+2.95})&  &({\color[HTML]{349C07}+4.96} / {\color[HTML]{349C07}+11.55})&  &({\color[HTML]{C82F09}-5.03} / {\color[HTML]{349C07}+4.70})&  &({\color[HTML]{C82F09}-0.87} / {\color[HTML]{C82F09}-4.00})&  \\[2pt]
\multicolumn{1}{c|}{} &  & ResNet & \multicolumn{1}{l|}{}       &  & 85.71	/ 86.35   &  &  90.64 / 91.95   &  &  89.50 / 92.00   &  & \textbf{91.37	/ 93.30} &  \\
\multicolumn{1}{c|}{} &  &              & \multicolumn{1}{l|}{} &  &({\color[HTML]{349C07}+1.13} / {\color[HTML]{349C07}+0.25})&  &({\color[HTML]{349C07}+3.75} / {\color[HTML]{349C07}+2.50})&  &({\color[HTML]{349C07}+0.46} / {\color[HTML]{349C07}+0.30})&  &({\color[HTML]{349C07}+2.05} / {\color[HTML]{349C07}+1.95})&  \\[2pt]
\multicolumn{1}{c|}{} &  & SqueezeNet & \multicolumn{1}{l|}{}   &  & 77.23	/ 80.40  &  &  90.38	/ 92.25   &  &  89.19	/ 92.00   &  & \textbf{91.56	/ 94.35} &  \\
\multicolumn{1}{c|}{} &  &              & \multicolumn{1}{l|}{} &  &({\color[HTML]{349C07}+1.80} / {\color[HTML]{349C07}+0.65})&  &({\color[HTML]{349C07}+9.77} / {\color[HTML]{349C07}+7.05})&  &({\color[HTML]{349C07}+1.76} / {\color[HTML]{349C07}+1.75})&  &({\color[HTML]{349C07}+1.05} / {\color[HTML]{349C07}+2.05})&  \\[2pt]
\multicolumn{1}{c|}{} &  & VGG & \multicolumn{1}{l|}{}          &  & \textbf{85.64		/ 90.90}   &  &  28.38	/ 37.30 &  &  15.06	/ 23.25 &  & 23.10 / 28.15 &  \\
\multicolumn{1}{c|}{} &  &              & \multicolumn{1}{l|}{} &  &({\color[HTML]{C82F09}-4.25} / {\color[HTML]{349C07}+0.05})&  &({\color[HTML]{C82F09}-16.93} / {\color[HTML]{C82F09}-15.05})&  &({\color[HTML]{349C07}+1.94} / {\color[HTML]{349C07}+4.15})&  &({\color[HTML]{349C07}+2.28} / {\color[HTML]{C82F09}-3.15})&  \\\cmidrule{2-13}
\multicolumn{1}{c|}{} &  & Average & \multicolumn{1}{l|}{}      &  & 73.97		/ 82.46  &  &  81.29	/ \textbf{86.56}  &  &  78.25	/ 86.01   &  & \textbf{82.43}	/ 86.18 &  \\
\multicolumn{1}{c|}{} &  &              & \multicolumn{1}{l|}{} &  &({\color[HTML]{349C07}+2.25} / {\color[HTML]{349C07}+1.85})&  &({\color[HTML]{349C07}+5.48} / {\color[HTML]{349C07}+5.00})&  &({\color[HTML]{C82F09}-0.25} / {\color[HTML]{349C07}+1.24})&  &({\color[HTML]{349C07}+0.63} / {\color[HTML]{349C07}+0.45})&  \\ \bottomrule
\end{tabular}
\end{table*}

\begin{table*}
\renewcommand\bfdefault{b}
\footnotesize
\centering
\caption{\textbf{Effect of gating on FeatureFormer.} Results are reported on the NNEQ
dataset (out-of-domain), with average and best reported over 10 independent trials.}
\label{tab:gating_app}
\begin{tabular}{@{}cllllclcl@{}}
\toprule
Metric &  & Test Domain &  &  & \begin{tabular}[c]{@{}c@{}}Without Gating\\ (avg / best)\end{tabular} &  & \begin{tabular}[c]{@{}c@{}}With Gating\\ (avg / best)\end{tabular} &  \\ \midrule
\multicolumn{1}{c|}{\multirow{11}{*}{MAPE $\downarrow$}} &  & AlexNet & \multicolumn{1}{l|}{} &  & 9.26 / 7.12 &  & \textbf{7.29 / 5.42} &  \\
\multicolumn{1}{c|}{} &  & EfficientNet & \multicolumn{1}{l|}{} &  & 3.11 / 2.51 &  & \textbf{2.89 / 2.47} &  \\
\multicolumn{1}{c|}{} &  & GoogLeNet & \multicolumn{1}{l|}{} &  & \textbf{3.49 / 3.19} &  & 3.63 / 3.35 &  \\
\multicolumn{1}{c|}{} &  & MnasNet & \multicolumn{1}{l|}{} &  & 3.42 / \textbf{2.50} &  & \textbf{3.10} / 2.72 &  \\
\multicolumn{1}{c|}{} &  & MobileNetV2 & \multicolumn{1}{l|}{} &  & \textbf{2.73 / 2.12} &  & 3.04 / 2.70 &  \\
\multicolumn{1}{c|}{} &  & MobileNetV3 & \multicolumn{1}{l|}{} &  & \textbf{5.03 / 4.66} &  & 5.18 / 4.79 &  \\
\multicolumn{1}{c|}{} &  & NasBench201 & \multicolumn{1}{l|}{} &  & 9.66 / 7.69 &  & \textbf{8.94 / 7.45} &  \\
\multicolumn{1}{c|}{} &  & ResNet & \multicolumn{1}{l|}{} &  & 4.82	/ 4.49 &  & \textbf{4.65 / 4.36} &  \\
\multicolumn{1}{c|}{} &  & SqueezeNet & \multicolumn{1}{l|}{} &  & 5.03	/ 4.77 &  & \textbf{4.57 / 4.15} &  \\
\multicolumn{1}{c|}{} &  & VGG & \multicolumn{1}{l|}{} &  & 23.84 / 21.75 &  & \textbf{23.05 / 21.17} &  \\\cmidrule{2-9} 
\multicolumn{1}{c|}{} &  & Average & \multicolumn{1}{l|}{} &  & 7.04 / 6.08 &  & \textbf{6.63 / 5.86} &  \\ \midrule
\multicolumn{1}{c|}{\multirow{11}{*}{{Acc (10\%) $\uparrow$}}} &  & AlexNet & \multicolumn{1}{l|}{} &  & 61.73 / 74.35 &  & \textbf{73.52 / 85.80} &  \\
\multicolumn{1}{c|}{} &  & EfficientNet & \multicolumn{1}{l|}{} &  & 98.01 / 99.60 &  & \textbf{98.62 / 99.75} &  \\
\multicolumn{1}{c|}{} &  & GoogLeNet & \multicolumn{1}{l|}{} &  & \textbf{96.90	/ 98.20} &  & 96.62 / 97.95 &  \\
\multicolumn{1}{c|}{} &  & MnasNet & \multicolumn{1}{l|}{} &  & 98.40 / 99.60 &  & \textbf{98.76 / 99.75} &  \\
\multicolumn{1}{c|}{} &  & MobileNetV2 & \multicolumn{1}{l|}{} &  & \textbf{99.18 / 99.85} &  & 98.48 / 99.15 &  \\
\multicolumn{1}{c|}{} &  & MobileNetV3 & \multicolumn{1}{l|}{} &  & \textbf{88.65 / 91.05} &  & 87.67 / 89.70 &  \\
\multicolumn{1}{c|}{} &  & NasBench201 & \multicolumn{1}{l|}{} &  & 60.88 / 72.35 &  & \textbf{64.61 / 73.90} &  \\
\multicolumn{1}{c|}{} &  & ResNet & \multicolumn{1}{l|}{} &  & 90.18 / \textbf{93.55} &  & \textbf{91.37} / 93.30 &  \\
\multicolumn{1}{c|}{} &  & SqueezeNet & \multicolumn{1}{l|}{} &  & 88.59 / 90.30 &  & \textbf{91.56 / 94.35} &  \\
\multicolumn{1}{c|}{} &  & VGG & \multicolumn{1}{l|}{} &  & \textbf{24.09 / 29.85} &  & 23.10 / 28.15 &  \\ \cmidrule{2-9} 
\multicolumn{1}{c|}{} &  & Average & \multicolumn{1}{l|}{} &  & 80.66 / 84.87 &  & \textbf{82.43 / 86.18} &  \\ \bottomrule
\end{tabular}
\end{table*}

\begin{table*}
\renewcommand\bfdefault{b}
\footnotesize
\centering
\caption{\textbf{Effect of different masking schemes on FeatureFormer}. Results are reported on the NNEQ dataset (out-of-domain) over 3 independent trials, utilizing 4 attention heads in each configuration.}
\label{tab:app_mask}
\begin{tabular}{@{}cllllclclcl@{}}
\toprule
Metric &  & Test Domain &  &  & \begin{tabular}[c]{@{}c@{}}Global\\ (avg / best)\end{tabular} &  & \begin{tabular}[c]{@{}c@{}}$A + A^T + A^TA + AA^T$\\ (avg / best)\end{tabular} & & \begin{tabular}[c]{@{}c@{}}$A, A^T, A^TA, AA^T$\\ (avg / best)\end{tabular} \\ \midrule
\multicolumn{1}{c|}{\multirow{11}{*}{{MAPE $\downarrow$}}} &  & AlexNet & \multicolumn{1}{l|}{} &  & 8.62 / 7.61 &  & 6.49 / 5.60 &  & \textbf{5.67 / 5.42} \\
\multicolumn{1}{c|}{} &  & EfficientNet & \multicolumn{1}{l|}{} &  & 4.55 / 4.04 &  & 3.33 / 3.18 &  & \textbf{2.62 / 2.48} \\
\multicolumn{1}{c|}{} &  & GoogLeNet & \multicolumn{1}{l|}{} &  & 3.59 / 3.39 &  & 3.57 / 3.43 &  & \textbf{3.57 / 3.37} \\
\multicolumn{1}{c|}{} &  & MnasNet & \multicolumn{1}{l|}{} &  & 4.18 / 3.92 &  & 3.55 / 3.33 &  & \textbf{2.88 / 2.79} \\
\multicolumn{1}{c|}{} &  & MobileNetV2 & \multicolumn{1}{l|}{} &  & 3.76 / 3.61 &  & 3.32 / 3.24 &  & \textbf{3.08 / 3.03} \\
\multicolumn{1}{c|}{} &  & MobileNetV3 & \multicolumn{1}{l|}{} &  & 5.97 / 5.61 &  & \textbf{5.09} / 4.82 &  & 5.33 / \textbf{4.79} \\
\multicolumn{1}{c|}{} &  & NasBench201 & \multicolumn{1}{l|}{} &  & 21.31 / 20.56 &  & 12.23 / 10.84 &  & \textbf{8.89 / 7.62} \\
\multicolumn{1}{c|}{} &  & ResNet & \multicolumn{1}{l|}{} &  & 5.18 / 4.83 &  & 4.85 / 4.70 &  & \textbf{4.51 / 4.36} \\
\multicolumn{1}{c|}{} &  & SqueezeNet & \multicolumn{1}{l|}{} &  & 5.56 / 5.29 &  & 5.16 / 5.03 &  & \textbf{4.43 / 4.15} \\
\multicolumn{1}{c|}{} &  & VGG & \multicolumn{1}{l|}{} &  & 21.83 / 20.37 &  & \textbf{21.56 / 19.84} &  & 21.84 / 21.17 \\\cmidrule{2-10} 
\multicolumn{1}{c|}{} &  & Average & \multicolumn{1}{l|}{} &  & 8.45 / 7.92 &  & 6.92 / 6.40 &  & \textbf{6.28 / 5.92} \\ \midrule
\multicolumn{1}{c|}{\multirow{11}{*}{{Acc (10\%) $\uparrow$}}}  &  & AlexNet & \multicolumn{1}{l|}{} &  & 63.40 / 68.30 &  & 77.53 / 85.00 &  & \textbf{84.32 / 85.80} \\
\multicolumn{1}{c|}{} &  & EfficientNet & \multicolumn{1}{l|}{} &  & 91.90 / 94.60 &  & 98.17 / 98.90 &  & \textbf{99.42 / 99.65} \\
\multicolumn{1}{c|}{} &  & GoogLeNet & \multicolumn{1}{l|}{} &  & 96.78 / 97.75 &  & 96.97 / 97.80 &  & \textbf{97.05 / 97.95} \\
\multicolumn{1}{c|}{} &  & MnasNet & \multicolumn{1}{l|}{} &  & 94.28 / 96.55 &  & 97.23 / 97.50 &  & \textbf{99.12 / 99.45} \\
\multicolumn{1}{c|}{} &  & MobileNetV2 & \multicolumn{1}{l|}{} &  & 95.63 / 96.50 &  & 97.67 / 97.85 &  & \textbf{98.38 / 98.70} \\
\multicolumn{1}{c|}{} &  & MobileNetV3 & \multicolumn{1}{l|}{} &  & 83.37 / 86.45 &  & \textbf{88.43} / 89.20 &  & 86.58 / \textbf{89.70} \\
\multicolumn{1}{c|}{} &  & NasBench201 & \multicolumn{1}{l|}{} &  & 31.05 / 34.30 &  & 49.10 / 56.30 &  & \textbf{64.08 / 73.90} \\
\multicolumn{1}{c|}{} &  & ResNet & \multicolumn{1}{l|}{} &  & 87.47 / 90.25 &  & 90.53 / 91.25 &  & \textbf{92.43 / 93.10} \\
\multicolumn{1}{c|}{} &  & SqueezeNet & \multicolumn{1}{l|}{} &  & 84.87 / 86.65 &  & 87.23 / 88.40 &  & \textbf{92.57 / 94.35} \\
\multicolumn{1}{c|}{} &  & VGG & \multicolumn{1}{l|}{} &  & \textbf{29.87} / 33.50 &  & 28.47 / \textbf{34.50} &  & 27.20 / 28.15 \\\cmidrule{2-10} 
\multicolumn{1}{c|}{} &  & Average & \multicolumn{1}{l|}{} &  & 75.86 / 78.49 &  & 81.13 / 83.67 &  & \textbf{84.12 / 86.08} \\ \bottomrule
\end{tabular}
\end{table*}

\begin{table*}
\setlength{\tabcolsep}{4pt}
\renewcommand\bfdefault{b}
\ssmall
\centering
\caption{\textbf{Importance of $F_{op}$, $F_{attr}$ , and $F_{comp}$ to FeatureFormer on NNLQ \citep{liu2022nnlqp} (out-of-domain).} Results are reported over 3 independent trials. It should be noted that all results on this page correspond to the same table which has been split in half due to space constraints.}
\label{tab:app_enc}
\begin{tabular}{@{}c|l|ccccccc@{}}
\toprule
Metric & Test Domain 
& \begin{tabular}[c]{@{}c@{}} (1) $F_{comp}$ \\ (avg / best)\end{tabular}
& \begin{tabular}[c]{@{}c@{}} (2) $F_{attr}$\\ (avg / best)\end{tabular}
& \begin{tabular}[c]{@{}c@{}} (3) $F_{attr}$, $F_{comp}$\\ (avg / best)\end{tabular}
& \begin{tabular}[c]{@{}c@{}} (4) $F_{op}$\\ (avg / best)\end{tabular}
& \begin{tabular}[c]{@{}c@{}} (5) $F_{op}$, $F_{comp}$ \\ (avg / best)\end{tabular}
& \begin{tabular}[c]{@{}c@{}} (6) $F_{op}$, $F_{attr}$ \\ (avg / best)\end{tabular}
& \begin{tabular}[c]{@{}c@{}} (7) $F_{op}$, $F_{attr}$, $F_{comp}$\\ (avg / best)\end{tabular} \\
\midrule

\multicolumn{1}{c|}{\multirow{11}{*}{\rotatebox{90}{MAPE $\downarrow$}}}
& AlexNet        & 20.68	/  20.03 & 11.67	/ 11.35 & 10.39	 / 9.87 & 18.96	/ 17.55 & 16.97	/ 15.87 & 11.82	/ 11.43 & \textbf{10.34	/ 9.51} \\
& EfficientNet   & 21.08	/ 20.87 & 5.28	/ 4.63 & 5.36	/ 4.87 & 21.73	/ 21.33 & 19.29	/ 18.69 & 4.76	/ 4.56 & \textbf{4.23	/ 3.63} \\
& GoogLeNet      & 18.96	/ 18.69 & 5.25	/ 4.92 & 5.05	/ 4.97 & 20.58	/ 18.72 & 18.57	/ 18.28 & 4.94	/ \textbf{4.79} & \textbf{4.84}	/ 4.82 \\
& MnasNet        & 15.71	/ 15.47 & 2.01	/ 1.97 & 1.93	/ 1.85 & 21.07	/ 20.99 & 16.72	/ 16.48 & 2.04	/ 1.89 & \textbf{1.88	/ 1.83} \\
& MobileNetV2    & 14.63	/ 14.33 & 2.96	/ 2.48 & 2.94	/ \textbf{2.41} & 19.95	/ 19.87 & 14.04	/ 13.92 & 4.05	/ 3.29 & \textbf{2.88}	/ 2.50 \\
& MobileNetV3    & 29.12	/ 27.86 & \textbf{7.85	/ 7.68} & 7.95	/ 7.78 & 27.86	/ 27.71 & 28.75	/ 27.50 & 8.33	/ 7.97 & 8.09/ 	8.03 \\
& NasBench201    & 13.08	/ 5.58 & 9.36	/ 8.31 & \textbf{5.27	/ 4.38} & 80.52	/ 38.87 & 8.08	/ 5.55 & 7.27	/ 4.03 & 7.19	/ 4.98 \\
& ResNet         & 17.14	/ 16.28 & 10.82	/ 10.35 & 8.18	/ \textbf{7.13} & 16.52	/ 15.94 & 17.49	/ 16.96 & 8.93	/ 8.56 & \textbf{7.56}	/ 7.26 \\
& SqueezeNet     & 17.22	/ 17.01 & 8.31	/ 7.65 & 6.44	/ 6.08 & 15.79	/ 15.57 & 17.85	/ 17.57 & 7.11	/ 6.19 & \textbf{6.37	/ 6.02}  \\
& VGG            & 30.04	/ 28.82 & 18.32	/ 17.29 & 17.88	/ 17.70 & 30.80	/ 30.40 & 27.68	/ 25.60 & \textbf{17.79	/ 17.04} & 18.58	/ 17.30 \\
\cmidrule{2-9}
& Average        & 19.77	/ 18.49 & 8.18	/ 7.66 & \textbf{7.14} / 	6.70 & 27.38	/ 22.69 & 18.54	/ 17.64 & 7.70	/ 6.97 & 7.20	/ \textbf{6.59} \\

\midrule

\multicolumn{1}{c|}{\multirow{11}{*}{\rotatebox{90}{Acc (10\%) $\uparrow$}}}
& AlexNet        & 32.40	/ 33.50 & 54.47	/ 55.90 & \textbf{59.67}	/ 63.15 & 33.92	/ 34.65 & 36.68	/ 38.05 & 55.72	/ 56.60 & 59.27	/ \textbf{64.40} \\
& EfficientNet   & 27.93	/ 28.80 & 87.50	/ 91.50 & 87.43	/ 90.25 & 26.62	/ 27.50 & 31.03	/ 32.70 & 90.15	/ 90.90 & \textbf{93.10	/ 95.95} \\
& GoogLeNet      & 29.67	/ 30.90 & 87.38	/ 89.50 & 87.87	/ 88.60 & 26.97	/ 30.05 & 30.68	/ 32.55 & 88.98	/ 89.85 & \textbf{90.28	/ 90.55} \\
& MnasNet        & 41.97	/ 42.50 & 99.47	/ 99.70 & \textbf{99.72	/ 99.85} & 27.17	/ 27.95 & 38.38	/ 40.15 & 99.45	/ 99.65 & 99.70	/ 99.75 \\
& MobileNetV2    & 42.65	/ 43.10 & 98.15	/ \textbf{99.70} & \textbf{97.87}	/ 99.30 & 29.68	/ 29.95 &  44.52	/ 44.90 & 92.68	/ 97.55 & 97.57	/ 99.55 \\
& MobileNetV3    & 17.83	/ 19.75 & \textbf{74.22	/ 76.05} & 74.18	/ 75.25 & 19.25	/ 20.15 & 18.65	/ 19.85 & 71.90	/ 74.30 & 72.43	/ 73.95 \\
& NasBench201    & 53.42	/ 84.20 & 62.82	/ 68.60 & \textbf{88.17}	/ 93.20 & 15.00	/ 25.30 & 70.12	/ 84.15 & 74.23	/ \textbf{95.40} & 74.60	/ 90.25 \\
& ResNet         & 35.10	/ 38.60 & 51.58	/ 57.60 & 67.68	/ \textbf{74.75} & 35.17	/ 37.75 & 34.80	/ 37.45 & 61.75	/ 64.20 & \textbf{71.93}	/ 73.35 \\
& SqueezeNet     & 38.35	/ 39.10 & 66.90	/ 71.50 & \textbf{79.17}	/ 81.25 & 40.48	/ 41.85 & 36.43	/ 38.05 & 74.13	/ 80.95 & 79.10	/ \textbf{82.00} \\
& VGG            & 19.98	/ 21.85 & 33.97	/ 38.95 & \textbf{36.20}	/ 36.80 & 21.02	/ 22.00 & 21.87	/ 26.00 & 33.62	/ 34.20 & 36.05	/ \textbf{39.25} \\
\cmidrule{2-9}
& Average        & 33.93	/ 38.23 & 71.65	/ 74.90 & \textbf{77.80}	/ 80.24 & 27.53	/ 29.72 & 36.32	/ 39.39 & 74.26	/ 78.36 & 77.40	/ \textbf{80.90} \\

\bottomrule
\end{tabular}

\end{table*}

\begin{table*}
\renewcommand\bfdefault{b}
\ssmall
\centering
\caption{\textbf{Results of different makeups of $F_{comp}$ on FeatureFormer on NNLQ \citep{liu2022nnlqp} (out-of-domain).} Memory (abbreviated "Mem") refers to a proxy of memory operations within the model, consisting of the multiplied output shape + parameter count. Results are reported over 3 independent trials. It should be noted that all results on this page correspond to the same table which has been split in half due to space constraints.}
\label{tab:app_comp}
{
\setlength{\tabcolsep}{4pt}
\begin{tabular}{@{}c|l|cccccccc@{}}
\toprule
Metric & Test Domain 
& \begin{tabular}[c]{@{}c@{}} (1) No $F_{comp}$ \\ (avg / best)\end{tabular}
& \begin{tabular}[c]{@{}c@{}} (2) Mem\\ (avg / best)\end{tabular}
& \begin{tabular}[c]{@{}c@{}} (3) Params\\ (avg / best)\end{tabular}
& \begin{tabular}[c]{@{}c@{}} (4) Params \\ + Mem\\ (avg / best)\end{tabular}
& \begin{tabular}[c]{@{}c@{}} (5) FLOPs \\ (avg / best)\end{tabular}
& \begin{tabular}[c]{@{}c@{}} (6) FLOPs \\ + Mem \\ (avg / best)\end{tabular}
& \begin{tabular}[c]{@{}c@{}} (7) FLOPs \\ + Params \\ (avg / best)\end{tabular}
& \begin{tabular}[c]{@{}c@{}} (8) FLOPs + \\ Params + Mem\\ (avg / best)\end{tabular}\\
\midrule

\multicolumn{1}{c|}{\multirow{11}{*}{\rotatebox{90}{MAPE $\downarrow$}}}
& AlexNet        & 11.82 / 11.43 & 10.50	/ 10.15 & 10.54	/ 10.19 & \textbf{10.26}	/ 9.70 & 11.44 / 11.25 & 10.74 / 10.04 & 10.81 / 10.27 & 10.34 / \textbf{9.51} \\
& EfficientNet   & 4.76	/ 4.56 & 4.75	/ 4.55 & 4.96	/ 3.79 & 4.83	/ 3.97 & 5.14 /	4.01 & 4.36	/ 4.18 & 4.54	/ 3.67 & \textbf{4.23	/ 3.63} \\
& GoogLeNet      & 4.94	/ 4.79 & 4.88	/ 4.86 & 4.87	/ 4.77 & 4.87	/ 4.81 & 4.83	/ 4.72 & 5.25	/ 4.90 & \textbf{4.80	/ 4.64} & 4.84 / 	4.82 \\
& MnasNet        & 2.04	/ 1.89 & 1.97	 / 1.94 & 1.90	/ \textbf{1.82} & 1.97	/ 1.89 & 1.93	/ 1.86 & 1.89	/ 1.85 & \textbf{1.85}	/ 1.84 & 1.88	/ 1.83 \\
& MobileNetV2    & 4.05	/ 3.29 & 4.07	/ 3.61 & 3.89	/ 3.30 & 2.81	/ 2.66 & 3.18	/ 2.28 & 3.36	/ 2.93 & \textbf{2.69	/ 2.50} & 2.88	/ 2.50 \\
& MobileNetV3    & 8.33	/ 7.97 & 8.69	/ 8.47 & 8.88	/ 8.63 & 8.36	/ 7.90 & \textbf{7.68	/ 7.25} & 7.89	/ 7.72 & 7.90	/ 7.60 & 8.09	/ 8.03 \\
& NasBench201    & 7.27	/ 4.03 & \textbf{5.76	/ 4.67} & 7.90	/ 5.96 & 7.61	/ 5.75 & 6.66	/ 5.49 & 8.15	/ 7.08 & 7.70	/ 7.13 & 7.19	/ 4.98 \\
& ResNet         & 8.93	/ 8.56 & 8.45	/ 7.55 & 7.64	/ \textbf{6.97} & 8.17	/ 7.11 & 7.75	/ 7.00 & 7.90	/ 7.18 & 8.08	/ 7.71 & \textbf{7.56}	/ 7.26 \\
& SqueezeNet     & 7.11	/ 6.19 & 6.23	/ 5.90 & 6.22	/ \textbf{5.16} & 7.37	/ 6.68 & 8.71	/ 8.11 & 6.57	/ 5.78 & \textbf{5.87}	/ 5.45 & 6.37	/ 6.02 \\
& VGG            & \textbf{17.79 / 17.04} & 19.26	/ 19.10 & 19.17	/ 18.14 & 19.48	/ 19.10 & 18.18	/ 17.33 & 18.56	/ 18.32 & 18.04	/ 17.45 & 18.58	/ 17.30 \\
\cmidrule{2-10}
& Average        & 7.70	/ 6.97 & 7.46	/ 7.08 & 7.60	/ 6.87 & 7.57	/ 6.96 & 7.55	/ 6.93 & 7.47	/ 7.00 & 7.23	/ 6.82 & \textbf{7.20	/ 6.59} \\

\midrule

\multicolumn{1}{c|}{\multirow{11}{*}{\rotatebox{90}{Acc (10\%) $\uparrow$}}}
& AlexNet        & 55.72	/ 56.60 & 59.37	/ 61.60 & 59.60	/ 60.90 & \textbf{60.45}	/ 63.05 & 55.33	/ 55.55 & 57.73	/ 61.80 & 58.43	/ 59.55 & 59.27	/ \textbf{64.40} \\
& EfficientNet   & 90.15	/ 90.90 & 89.78	/ 90.90 & 89.35	/ 95.80 & 89.32	/ 94.35 & 87.95	/ 93.50 & 92.30	/ 93.15 & 91.38	/ 95.90 & \textbf{93.10	/ 95.95} \\
& GoogLeNet      & 88.98	/ 89.85 & 89.55	/ 90.00 & 89.38	/ 89.95 & 89.23	/ 89.90 & 88.92	/ 90.50 & 87.53	/ 89.25 & 89.87	/ \textbf{90.60} & \textbf{90.28}	/ 90.55 \\
& MnasNet        & 99.45	/ 99.65 & 99.57	/ 99.75 & 99.57	/ 99.75 & 99.50	/ 99.55 & 99.82	/ 99.90 & 99.78	/ 99.90 & \textbf{99.83	/ 99.90} & 99.70	/ 99.75 \\
& MobileNetV2    & 92.68	/ 97.55 & 90.58	/ 92.15 & 93.37	/ 98.00 & \textbf{98.92}	/ 99.40 & 95.80	/ \textbf{99.55} & 95.50	/ 98.45 & 98.62	/ 99.40 & 97.57	/ \textbf{99.55} \\
& MobileNetV3    & 71.90	/ 74.30 & 71.73	/ 72.05 & 72.33	/ 74.50 & 72.38	/ 74.05 & 74.15	/ \textbf{77.10} & 73.92	/ 74.85 & \textbf{74.40}	/ 76.00 & 72.43	/ 73.95 \\
& NasBench201    & 74.23	/ 95.40 & \textbf{83.27	/ 91.35} & 69.22	/ 82.75 & 71.53	/ 83.15 & 77.92	/ 86.85 & 68.83	/ 75.30 & 73.82	/ 77.70 & 74.60	/ 90.25 \\
& ResNet         & 61.75	/ 64.20 & 64.12	/ 70.05 & 71.23	/ 74.95 & 67.67	/ 75.70 & 70.40	/ \textbf{75.80} & 69.82	/ 75.25 & 67.35	/ 69.60 & \textbf{71.93}	/ 73.35 \\
& SqueezeNet     & 74.13	/ 80.95 & 80.68	/ 82.15 & 80.37	/ \textbf{87.40} & 72.87	/ 77.40 & 64.35	/ 69.20 & 78.77	/ 84.05 &\textbf{ 83.70}	/ 85.90 & 79.10	/ 82.00 \\
& VGG            & 33.62	/ 34.20 & 30.27	/ 31.95 & 33.12	/ 35.10 & 31.15	/ 32.60 & 34.02	/ 35.80 & 32.55	/ 34.50 & \textbf{36.58}	/ 38.90 & 36.05	/ \textbf{39.25} \\
\cmidrule{2-10}
& Average        & 74.26 / 78.36 & 75.89	/ 78.20 & 75.75	/ 79.91 & 75.30	/ 78.92 & 74.87	/ 78.38 & 75.67	/ 78.65 & 77.40	/ 79.35 & \textbf{77.40	/ 80.90}\\

\bottomrule
\end{tabular}
}

\end{table*}


\end{document}